\documentclass[letterpaper,10pt,conference]{ieeeconf}  

\IEEEoverridecommandlockouts                             
\PassOptionsToPackage{export}{adjustbox}
\usepackage{definitions}
\usepackage[T1]{fontenc}
\usepackage{tikz}
\usepackage{flushend}

\usepackage{dsfont}
\usepackage[cal=pxtx,scr=boondox]{mathalpha}
\usepackage{bm}
\usepackage{yhmath}
\usepackage{adjustbox}
\usepackage{mathtools}
\usepackage{amsmath}
\usepackage{amssymb} 
\usepackage[linesnumbered,ruled,vlined]{algorithm2e}
\usepackage{url}
\usepackage{graphics}
\usepackage{graphicx}
\usepackage{epsfig}
\usepackage{textcomp}
\usepackage{cite}
\usepackage{bm}
\usepackage{subcaption}
\usepackage{caption}
\usepackage{booktabs}
\usepackage{tabularx}
\usepackage{tabularray}
\usepackage[english]{babel}
\usepackage[utf8]{inputenc}
\usepackage{theorem}
\usepackage{pifont}
\usepackage{cuted}
\usepackage[dvipsnames]{colortbl,xcolor}
\colorlet{tablerowgray}{black!8} % Alternating table-body rows.
\usepackage{multirow}
\usepackage{makecell}
\usepackage{float}
\usepackage{siunitx}
\usepackage{gensymb}
\usepackage{stmaryrd}

\let\labelindent\relax
\usepackage{enumitem}

\newcommand{\ind}{\mathds{1}}

\newcommand{\bestresult}[1]{\mathbf{#1}}
\newcommand{\secondresult}[1]{\underline{#1}}

\title{\LARGE \bf JEPLO: Joint-Embedding Predictive Learning for \\LiDAR-Based Legged Locomotion}

\author{Qihao~Yuan, Yixuan Qiu, Ziyu~Cao, Ming~Cao, and Kailai~Li%
\thanks{Qihao Yuan, Yixuan Qiu, and Ming Cao are with the Faculty of Science and Engineering, University of Groningen, The Netherlands. Ziyu Cao is with the Department of Electrical Engineering, Linköping University, Sweden. Kailai Li was with the Faculty of Science and Engineering, University of Groningen, and is now with the Department of Electrical Engineering, Linköping University, Sweden.
E-mails: \tt{\{qihao.yuan, yixuan.qiu, m.cao\}@rug.nl},\tt{\{ziyu.cao, kailai.li\}@liu.se}}
}
     
\begin{document}
	
	\maketitle
	\thispagestyle{empty}
	\pagestyle{empty}
	
	\begin{abstract}
        Light detection and ranging (LiDAR) remains less explored than RGB-D sensing for perceptive legged locomotion, and existing LiDAR-based approaches often rely on explicit mapping. We present JEPLO (Joint-Embedding Predictive learning for legged LOcomotion), a single-stage learning framework for mapping-free, LiDAR-based perceptive locomotion for legged robots. We introduce a proprio-exteroceptive JEPA (PE-JEPA) world model to learn predictive egocentric terrain representations from onboard observations, including raw LiDAR scans. A concurrent JEPA-teacher-student (CJTS) pipeline is further proposed to train a locomotion policy informed by JEPA latent representations in simulation using deep reinforcement learning with a simple reward formulation. The framework achieves successful sim-to-real transfer, enabling omnidirectional traversal of diverse terrains, including long staircases and high boxes, with lightweight onboard computation. Evaluations demonstrate greater robustness than existing perceptive locomotion frameworks, particularly under degraded perception caused by occlusion, sparsity and noise. Further analysis validates JEPLO’s ability to retain task-relevant information under these challenging conditions. We open-source our implementation, experimental datasets, and hardware setup designs at \url{https://github.com/ASIG-X/JEPLO}.
	\end{abstract}
	
	\section{Introduction}\label{sec:intro}
    Perceptive locomotion enables legged robots to navigate complex, unstructured environments, supporting a wide range of applications, including exploration, search and rescue, and industrial inspection~\cite{rudin2025parkourinthewild,hoeller2024anymalparkour,miki2022inthewild}. Recent advances in deep reinforcement learning (RL) have substantially improved the ability of legged robots to traverse challenging terrain in real-world settings~\cite{cheng2024extreme,luo2024pie,lai2025wmp}. Such traversal relies on adaptive locomotion behaviors informed by egocentric perception. Learning task-relevant terrain representations from raw sensory observations is therefore particularly important for learning robust locomotion policies.
        
    RGB-D cameras have been widely adopted for legged perceptive locomotion because they provide direct and structured depth measurements~\cite{agarwal2023egocentric,cheng2024extreme,lai2025wmp,luo2024pie}. In this setting, a common approach to RL-based policy training of perceptive legged skills is teacher-student distillation: a teacher is first trained using privileged terrain information, typically a height map, and a student then learns to imitate the teacher using depth images and proprioceptive observations~\cite{cheng2024extreme,agarwal2023egocentric}. This strategy also allows multiple specialized teacher policies to be distilled into a single deployable policy~\cite{rudin2025parkourinthewild}. However, conventional two-stage training fixes the teacher's behavior without accounting for the student's perceptual limitations during policy learning. The teacher may learn behaviors that are difficult to reproduce under the student's sensing constraints, such as occlusion and field-of-view (FoV) limits, hindering reliable imitation from onboard observations. RL-based student fine-tuning can mitigate this mismatch by allowing the policy to adapt to its own sensing conditions, but adds training complexity and potential instability~\cite{rudin2025parkourinthewild}.

    \begin{figure}[t!]
        \centering
        \captionsetup[subfigure]{skip=2pt}
        \setlength{\tabcolsep}{0pt}
    
        \begin{tabular}{c}
            \subcaptionbox{Indoor environment testing and hardware setup.%
                \label{fig:teaser_indoor}}{%
                \includegraphics[width=0.97\columnwidth]
                {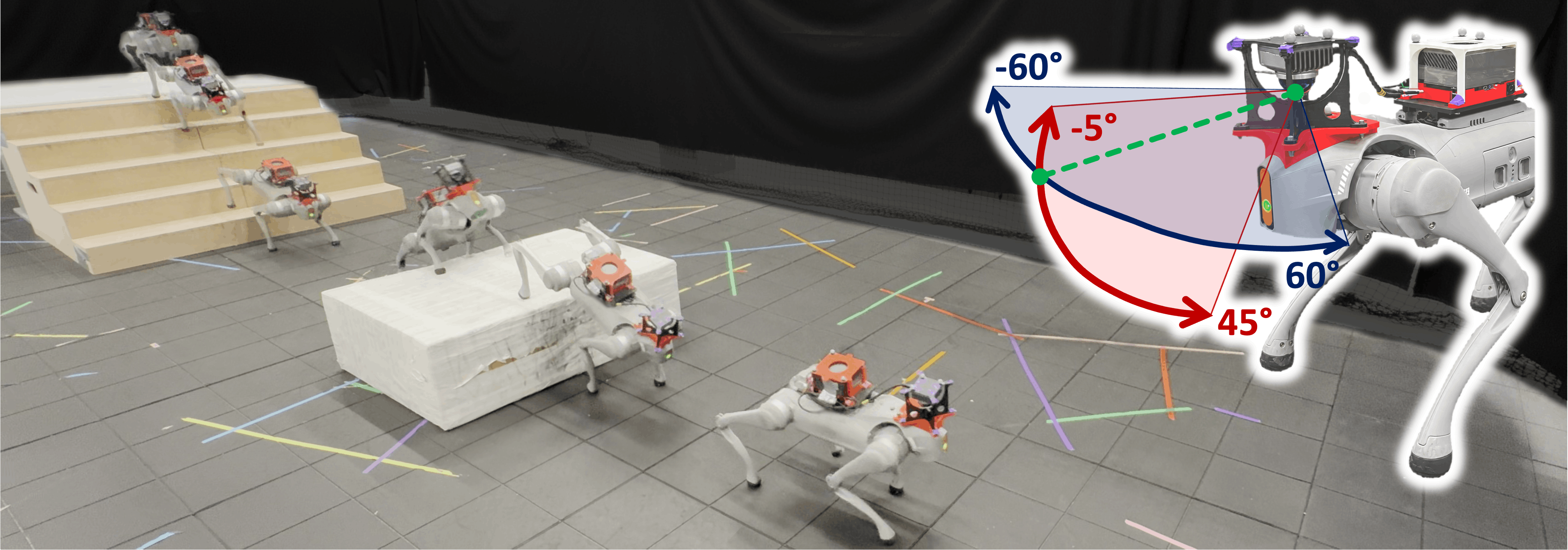}
            }
            \\[4pt]
            \subcaptionbox{Outdoor environment testing.%
                \label{fig:teaser_outdoor}}{%
                \includegraphics[width=0.97\columnwidth]
                {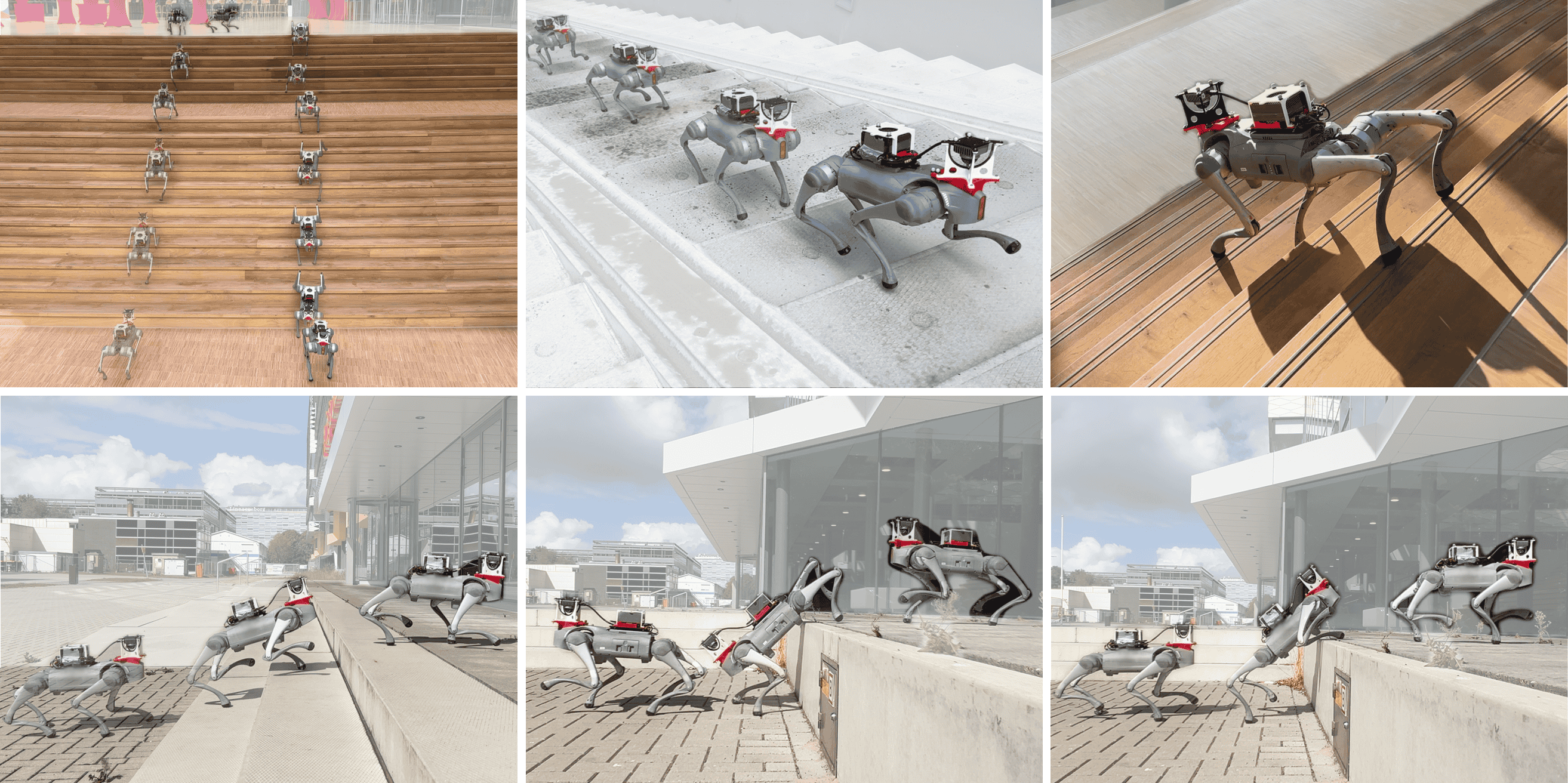}
            }
        \end{tabular}
    
        \caption{JEPLO enables mapping-free LiDAR-based perceptive locomotion
        onboard quadruped robots across diverse scenarios. Terrain height
        variations are \SIrange{18}{44}{\centi\meter}.}
        \label{fig:front}
        \vspace{-3mm}
    \end{figure}

    To learn complex motor skills in a single-stage manner, concurrent teacher-student training was proposed to combine distillation with RL through encoder-level supervision, allowing teacher and student encoders to improve continuously~\cite{wang2024cts}. Several methods similarly train perception modules and locomotion policies concurrently~\cite{luo2024pie,lai2025wmp}. The perception module learns through supervised or self-supervised objectives, providing explicit terrain estimates or latent features through height map reconstruction or world modeling~\cite{yu2025start,luo2024pie,lai2025wmp}. Meanwhile, the policy learns through RL, allowing perception and control to evolve together as the policy adapts to the information available from its sensors.

    Despite these advances, RGB-D cameras are susceptible to substantial noise and missing data under challenging light conditions such as darkness. LiDAR offers an alternative with accurate, wide-range spatial perception and greater robustness to lighting variations~\cite{yuan2025reasan}. However, individual LiDAR scans can provide incomplete terrain coverage, while aggregating scans without compensating for sensor motion can introduce severe geometric distortion. Many compact LiDAR sensors used on mobile robots employ non-repetitive, irregular scan patterns with sparse per-frame coverage. LiDAR-based locomotion systems therefore often rely on explicit mapping using odometry or SLAM to construct structured terrain maps~\cite{chen2025multilayer,hoeller2024anymalparkour,miki2022inthewild,cao2025resple}. Such systems incur additional computational overhead and system redundancy and can be sensitive to scene variations and perception degradation~\cite{miki2022elevation,rudin2025parkourinthewild}. These limitations motivate reactive perceptive locomotion via task-relevant terrain representation that remains robust to LiDAR-specific sensing artifacts.
    
    Latent-space predictive world models offer a promising approach to learning robust egocentric scene representations by capturing scene dynamics through prediction in an embedding space. In this context, Dreamer has been successfully combined with RL-based policy training for RGB-D-based perceptive locomotion~\cite{hafner2025dreamer,lai2025wmp}, reconstructing proprioceptive and depth observations while predicting future latent states to support locomotion control. However, real-world perceptive locomotion needs to contend with sensor noise, failures, and, for many LiDARs, irregular observation structures. Explicit scene reconstruction may force the model to reproduce these artifacts, consuming network capacity and increasing sensitivity to information irrelevant to control.

    Recently, JEPA has emerged as an approach to addressing this limitation by predicting in an embedding space without explicitly reconstructing observations, enabling abstraction from unnecessary pixel-level details~\cite{assran2023ijepa,assran2025vjepa2}. Its training objective allows representations to suppress unpredictable aspects of the input, such as random sensor artifacts. JEPA has demonstrated the ability to capture semantic and spatial structure in images, as well as physical state information and dynamics, with applications in video understanding, tabletop robotic manipulation, and quadrotor control~\cite{assran2023ijepa,maes2026leworldmodel,sun2026vlajepa,rao2026skyjepa}. However, to the best of our knowledge, joint-embedding predictive learning has not been investigated for perceptive locomotion in legged robots, where complex terrain contact interacts with evolving egocentric observations, particularly when using LiDARs with non-repetitive scan patterns.
    \begin{figure}[t!]
        \vspace{2mm}
        \centering
        \includegraphics[width=\linewidth]{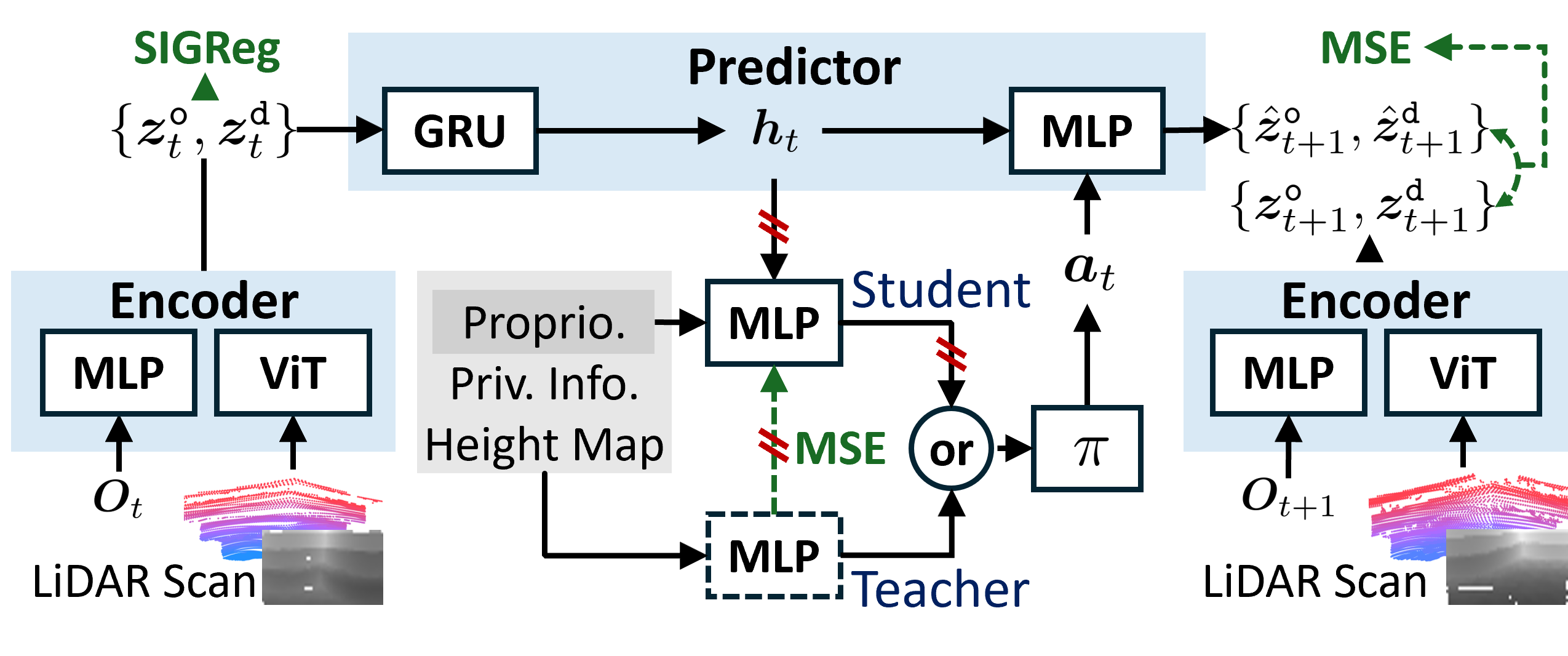}
        \caption{JEPLO pipeline. Slashes denote stop-gradient.}
        \label{fig:pipeline}
    \end{figure}
    
    \subsection*{Contributions}
    We propose {JEPLO}, a novel {J}oint-{E}mbedding {P}redictive learning framework for mapping-free, LiDAR-based {LO}comotion. A proprio-exteroceptive JEPA (PE-JEPA) world model is proposed for predictive modeling of egocentric terrain dynamics in embedding space for legged locomotion. The resulting latent features inform an RL-based locomotion policy trained in simulation through a single-stage concurrent JEPA-teacher-student (CJTS) pipeline. JEPLO enables traversal across diverse terrains, including long staircases and high boxes, and demonstrates greater robustness to perception artifacts than state-of-the-art perceptive locomotion approaches. Our contributions are summarized as follows.

    \begin{itemize}[leftmargin=*]
        \item We develop PE-JEPA, the first JEPA-based framework for mapping-free perceptive locomotion using non-repetitive LiDARs, jointly trained with an RL-based locomotion policy through a single-stage CJTS pipeline. Successful sim-to-real transfer enables traversal of challenging terrain with low onboard computational demands.
        \item Evaluations show that JEPLO outperforms state-of-the-art baselines under occlusion, sparsity and noise, despite irregular LiDAR scans. Latent analysis further validates that its representations retain task-relevant terrain information while reducing sensitivity to observation artifacts.
        \item We open-source the complete training and deployment implementation, the MuJoCo validation environment, experimental datasets, and hardware setup designs, supporting lightweight and robust legged autonomy.
    \end{itemize}

    \begin{figure*}[t!]
        \vspace{2mm}
        \centering
        \setlength{\tabcolsep}{1pt}
        \begin{tabular}{@{}cccccc@{}}
            \includegraphics[width=0.16\textwidth]{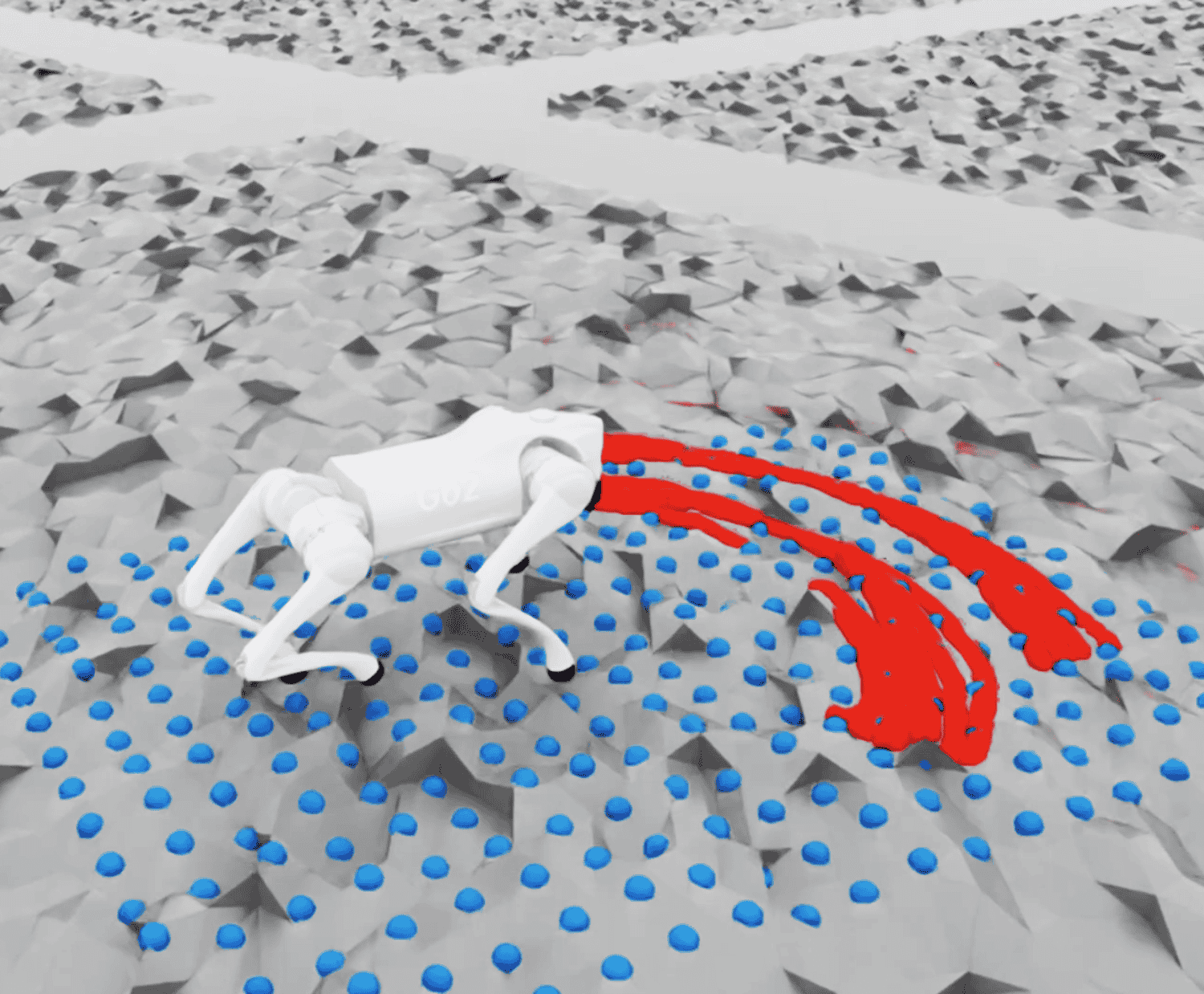}&
            \includegraphics[width=0.16\textwidth]{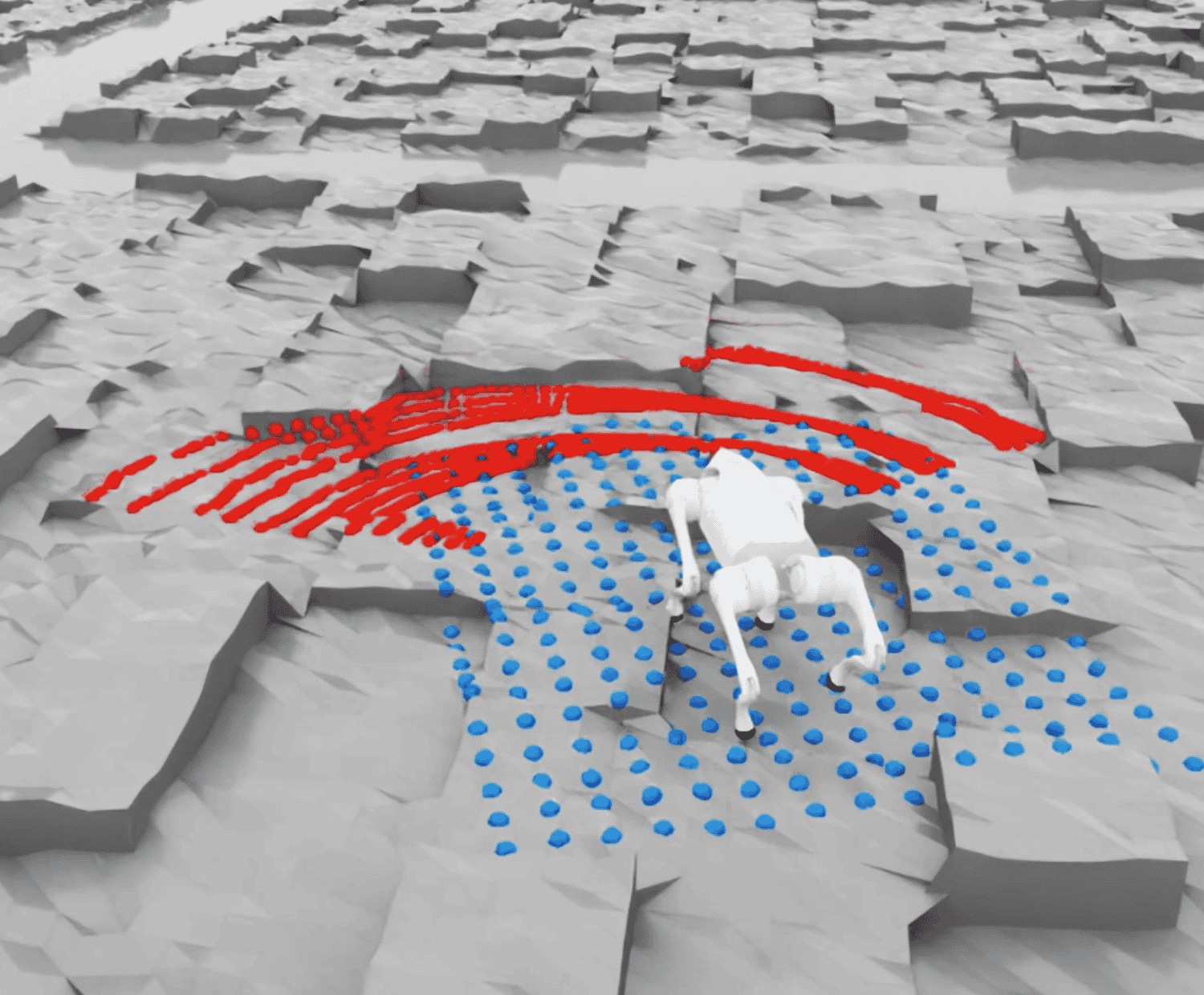}&
            \includegraphics[width=0.16\textwidth]{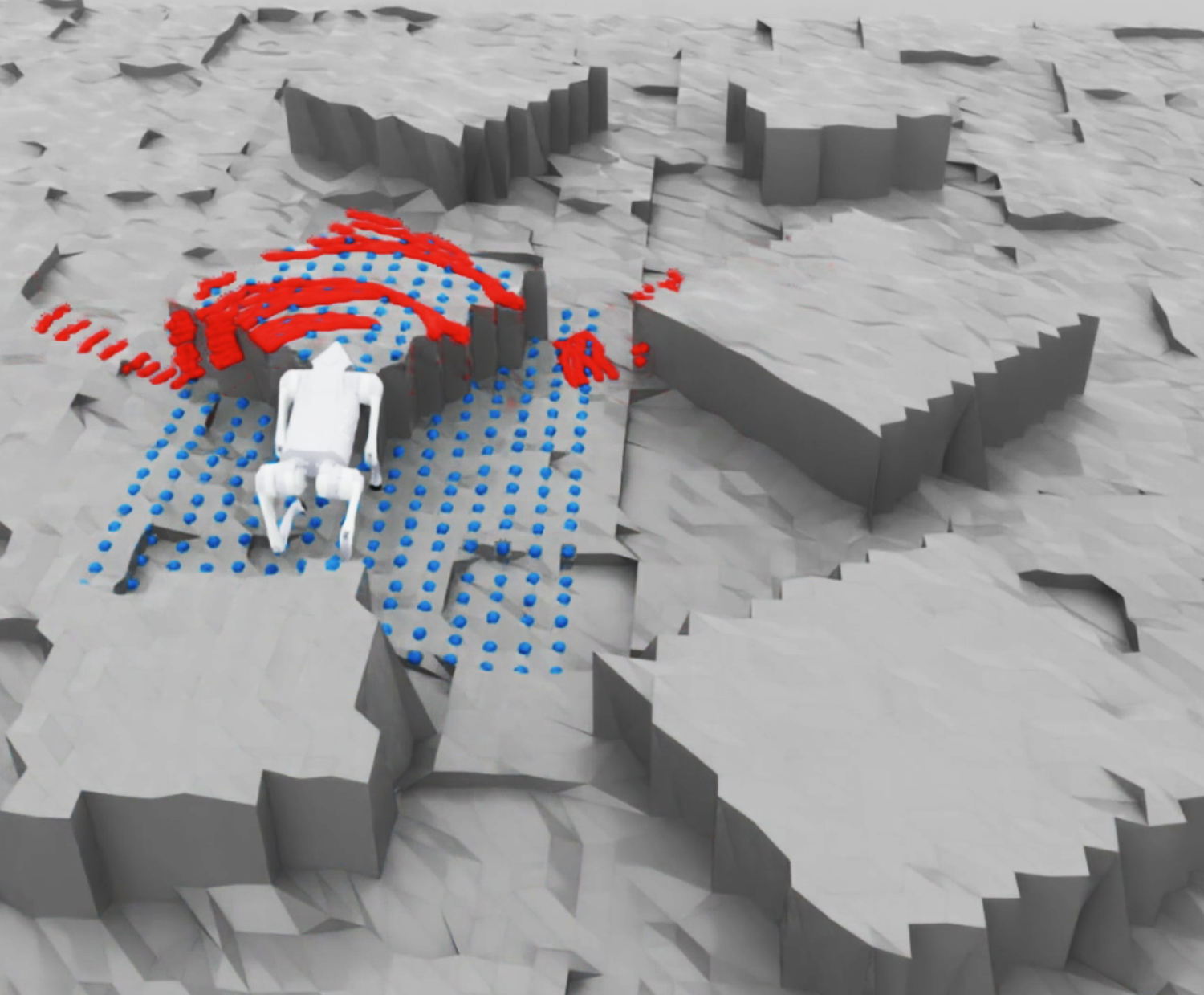}&
            \includegraphics[width=0.16\textwidth]{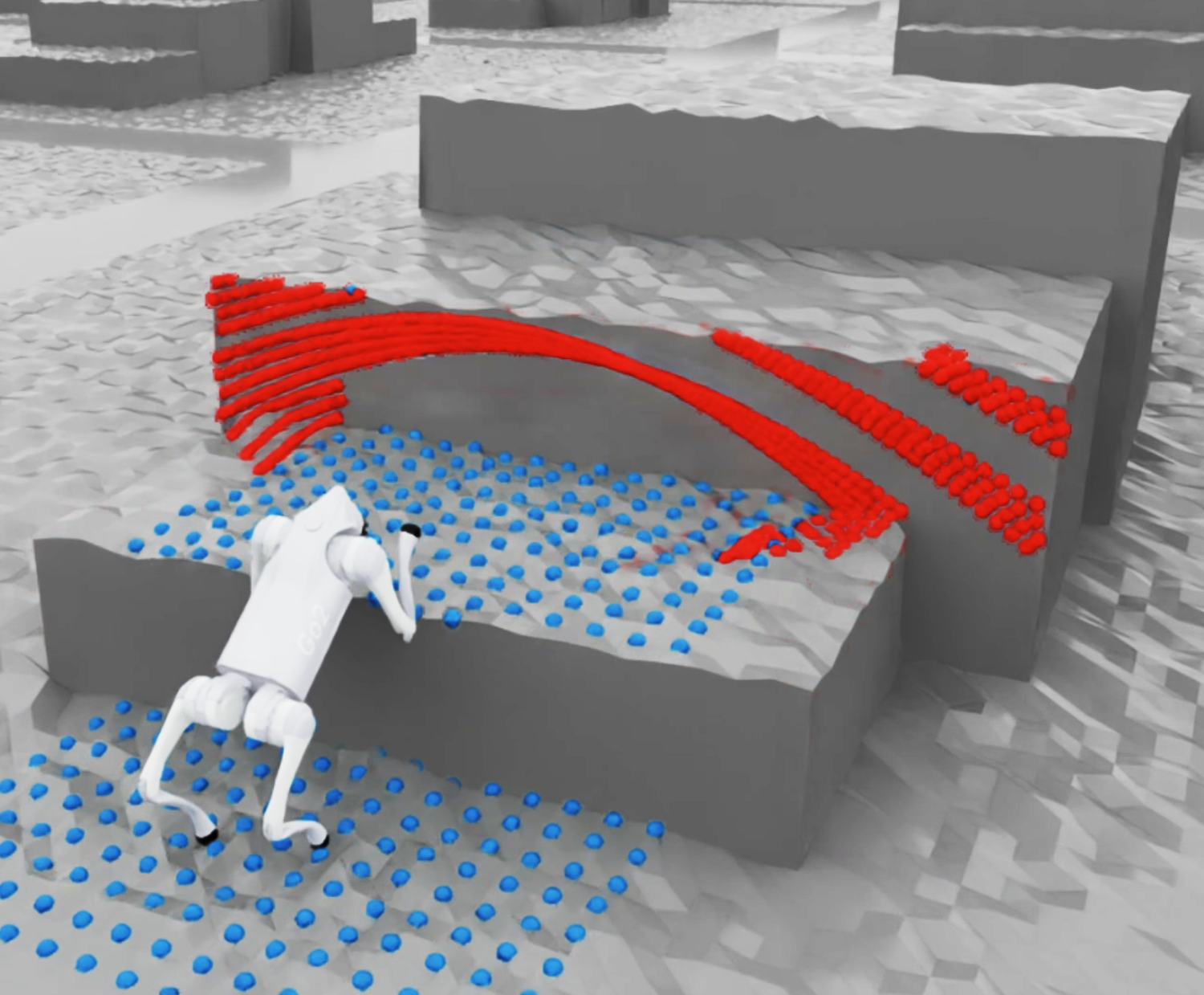}&
            \includegraphics[width=0.16\textwidth]{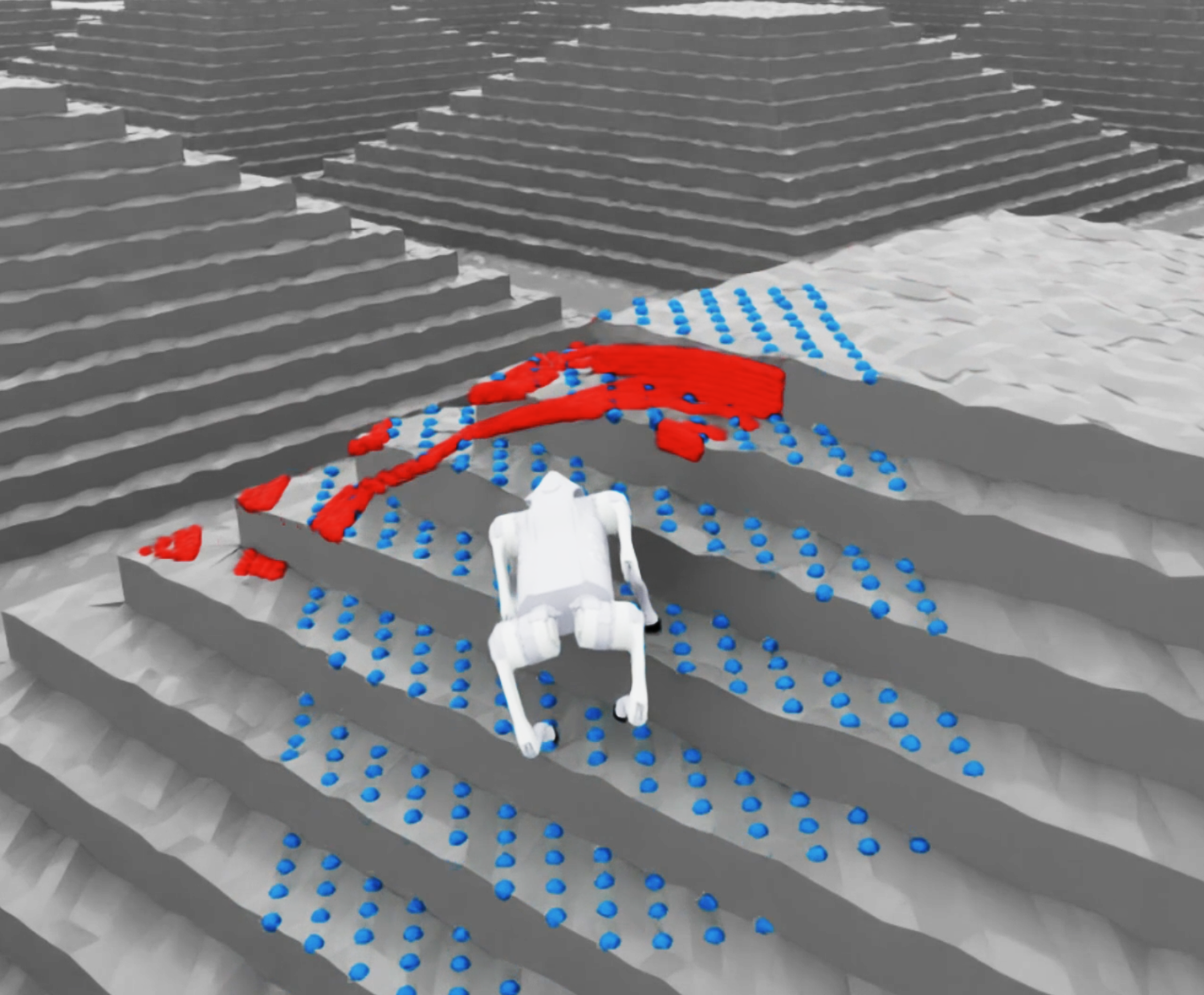}&
            \includegraphics[width=0.16\textwidth]{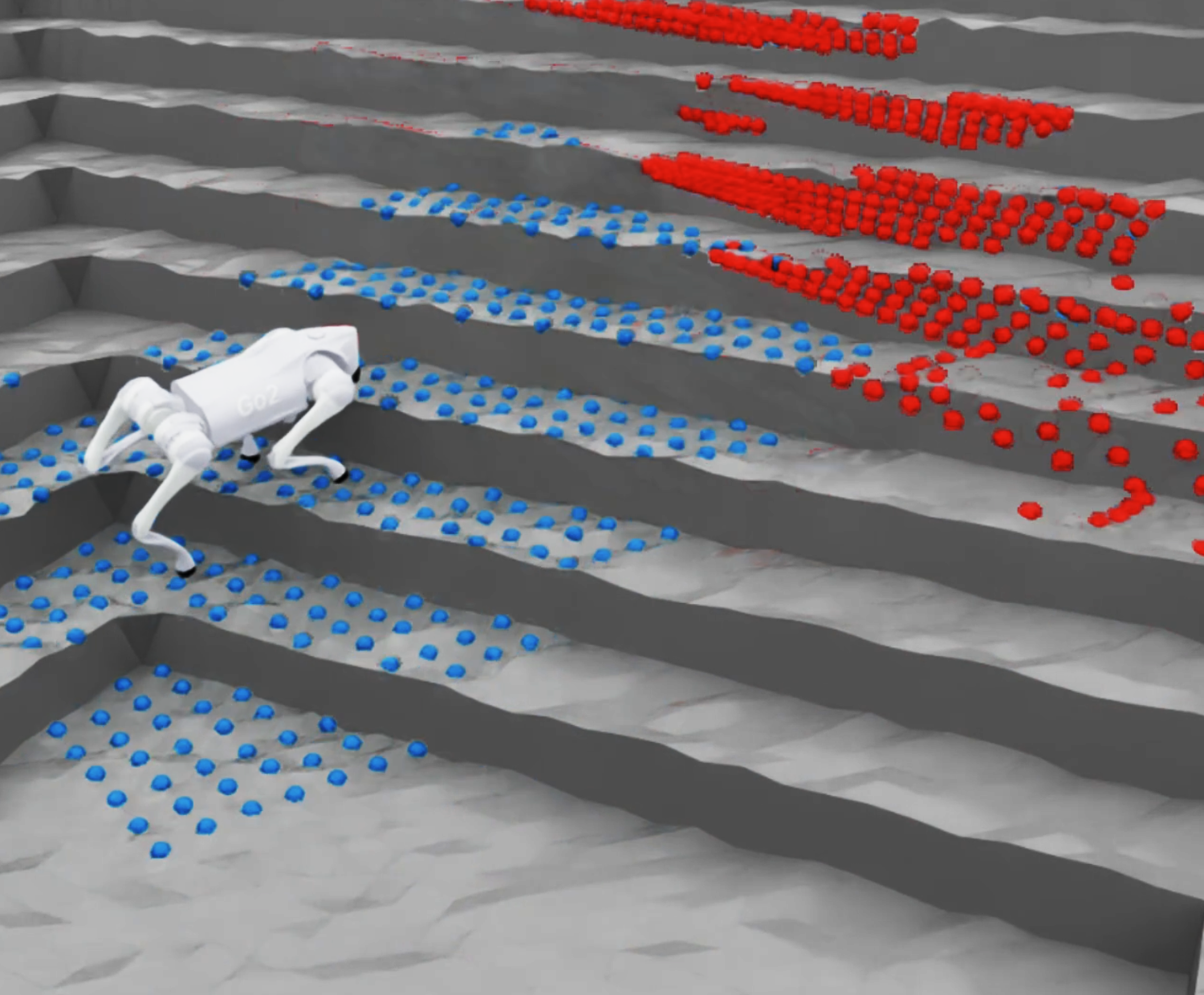}\\[2pt]
            (a) \ttt{Rough} & (b) \ttt{Discrete} & (c) \ttt{Boxes} & (d) \ttt{Steps} & (e) \ttt{Pyramid} & (f) \ttt{InvPyramid}
        \end{tabular}
        
        \caption{Training terrains. \ttt{Boxes} reach \SI{0.4}{\meter} in height. \ttt{Steps} have \SI{0.1}{}-\SI{0.5}{\meter} high increments. Stairs in (e-f) have \SI{0.04}{}-\SI{0.20}{\meter} rises and \SI{0.25}{}-\SI{0.35}{\meter} runs. Red and blue points show teacher height maps and student LiDAR scans, respectively.}
        \label{fig:training_terrains}
        % \vspace{-3mm}
    \end{figure*}
    
    \section{System Pipeline}\label{sec:system_pipeline}
    As shown in Fig.~\ref{fig:front}-(a), a Livox Mid-360 is mounted upside down on the front of a Unitree Go2 robot for terrain observation. We crop its horizontal field of view (FoV) to a forward-facing \SI{120}{\degree} sector and retain \SI{50}{\degree} of its vertical FoV, thereby excluding the region occluded by the computer mounted on the robot's back. The robot tracks the commanded planar velocity $\bm{v}^{\text{cmd}}_{\text{xy}}$ and yaw rate $\omega^{\text{cmd}}_{\text{z}}$, given LiDAR scans typically acquired at \SI{20}{\hertz}~\cite{hoeller2024anymalparkour}. We construct a depth image $\bm{d}_t$ by accumulating the latest five raw scans (without motion undistortion) and projecting the points onto a spherical grid covering the active FoV. With an angular resolution of \SI{2}{\degree}, the resulting $60\times25$ image stores the minimum point distance within each elevation-azimuth cell.
    
    As illustrated in Fig.~\ref{fig:pipeline}, the proposed JEPLO combines proprio-exteroceptive JEPA (PE-JEPA) world modeling with RL-based policy training through a concurrent JEPA-teacher-student (CJTS) pipeline with separate optimization objectives. PPO optimizes the actor, critic, and privileged teacher encoders, while the student encoder learns to imitate the teacher's embeddings. Concurrently, PE-JEPA learns predictive scene representations in latent space from sensor observations, providing the student encoder with terrain information for perceptive locomotion.

    The JEPA encoders comprise an MLP that maps the proprioceptive history $\bm{O}_t$ to a latent $\bm{z}_t^\text{o}\in\mathbb{R}^{32}$ and a Vision Transformer (ViT) that maps the depth history $\bm{D}_t$ to a latent $\bm{z}_t^\text{d}\in\mathbb{R}^{64}$~\cite{dosovitskiy2020vit}. The history $\bm{O}_t$ comprises a window of proprioceptive observations $\bm{o}_t\in\mathbb{R}^{45}$, each containing base roll-pitch rate $\bm{\omega}_{\text{xy}}$, yaw rate $\omega_{\text{z}}$, projected gravity, velocity commands $\bm{v}^{\text{cmd}}_{\text{xy}}$ and $\omega^{\text{cmd}}_{\text{z}}$, joint positions $\bm{\theta}$ and velocities $\dot{\bm{\theta}}$, and the previous action $\bm{a}_{t-1}$. The depth ViT uses $8\times4$ patches, six layers, four attention heads, and a 128-D embedding. The JEPA predictor includes a GRU with hidden state $\bm{h}_t\in\mathbb{R}^{512}$ informing the student encoder. MLP prediction heads use this state and the policy action sequence to predict next-step latents. The MLP teacher encoder produces a 192-D embedding that serves as the imitation target for the MLP student encoder. All MLPs in JEPA and the teacher-student encoders have three layers with hidden dimensions of 512 and 256. JEPA runs at \SI{10}{\hertz}, while the policy runs at \SI{50}{\hertz}~\cite{lai2025wmp}.

    \begin{figure}[t!]
    \vspace{2mm}
    \centering
    \captionsetup[subfigure]{skip=2pt}
    \setlength{\fboxsep}{0pt}
    \setlength{\fboxrule}{0.3pt}
    \setlength{\tabcolsep}{2pt}
    \begin{tabular}{cc}
        \subcaptionbox{\ttt{Steps}}{%
            \fbox{\includegraphics[
                width=\dimexpr0.235\columnwidth-2\fboxrule\relax]
                {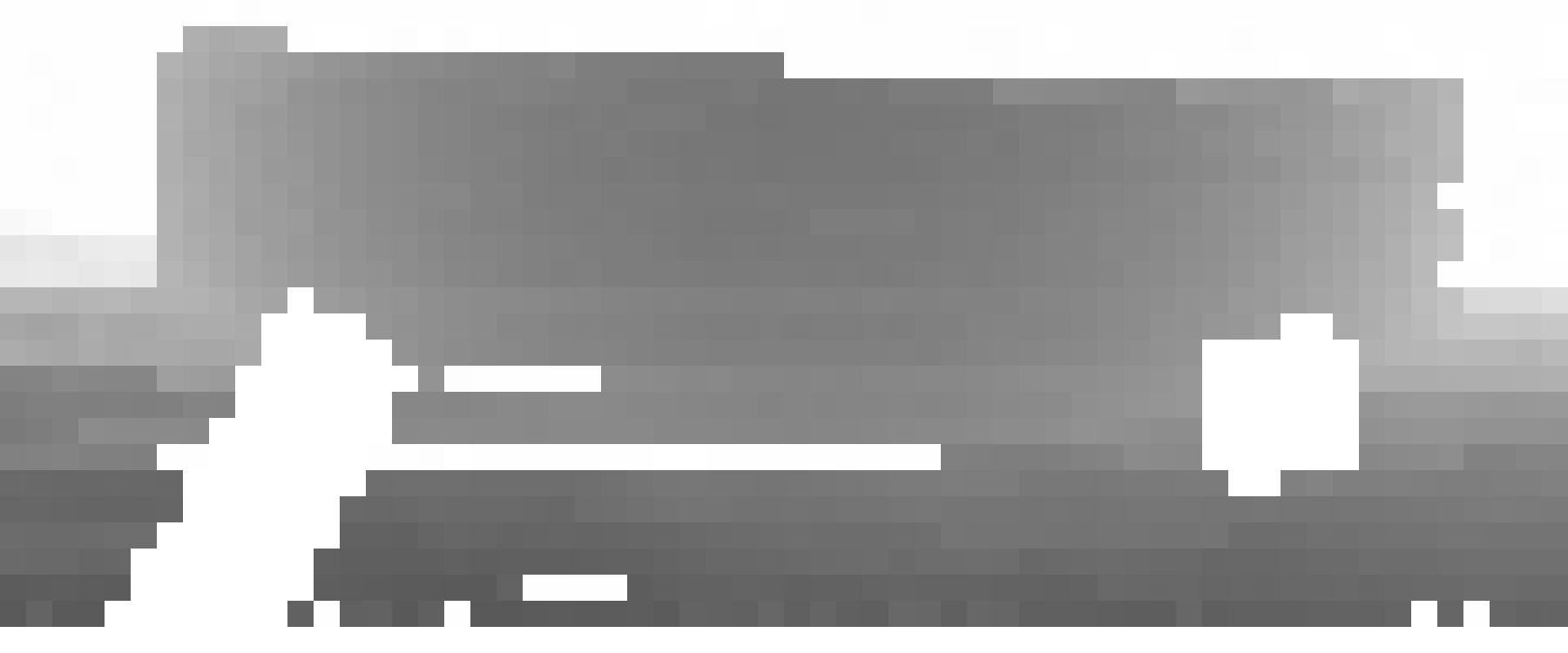}}%
            \hspace{1pt}%
            \fbox{\includegraphics[
                width=\dimexpr0.235\columnwidth-2\fboxrule\relax]
                {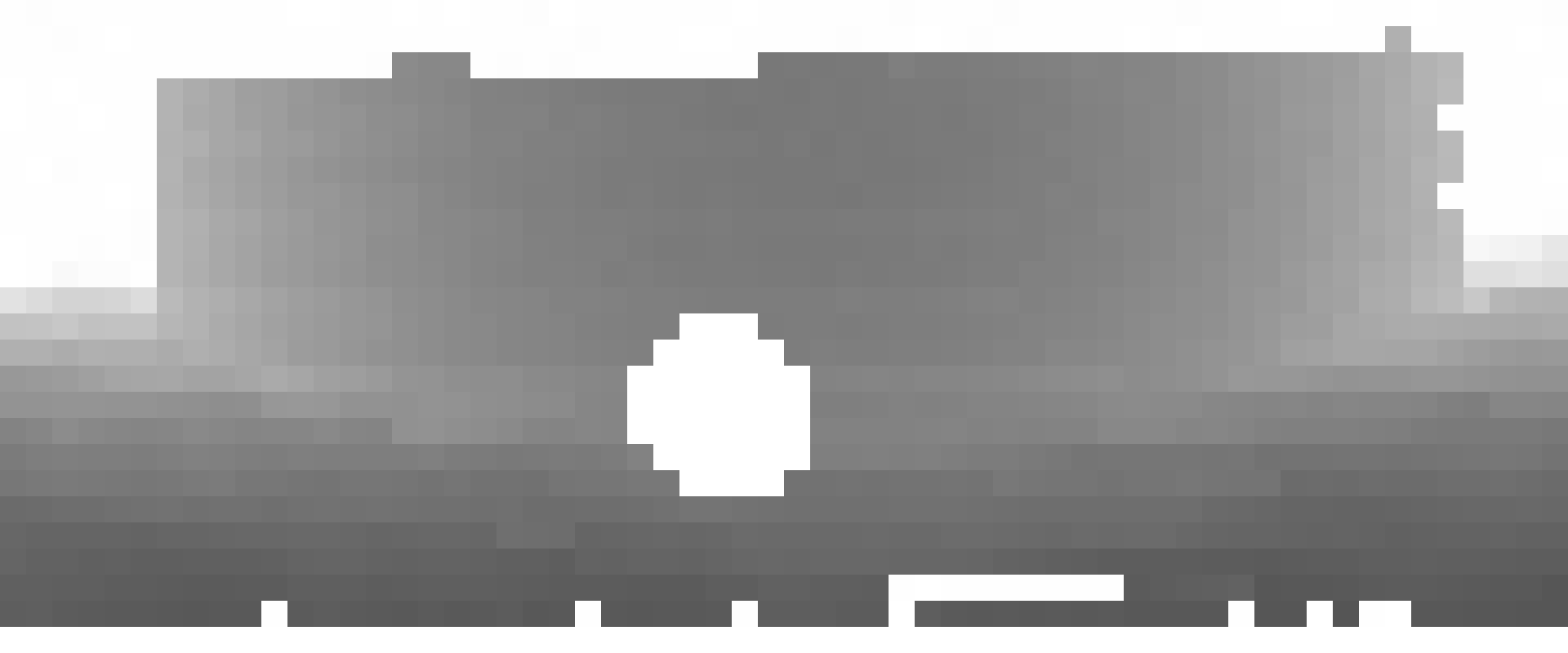}}%
        }
        &
        \subcaptionbox{\ttt{InvPyramid}}{%
            \fbox{\includegraphics[
                width=\dimexpr0.235\columnwidth-2\fboxrule\relax]
                {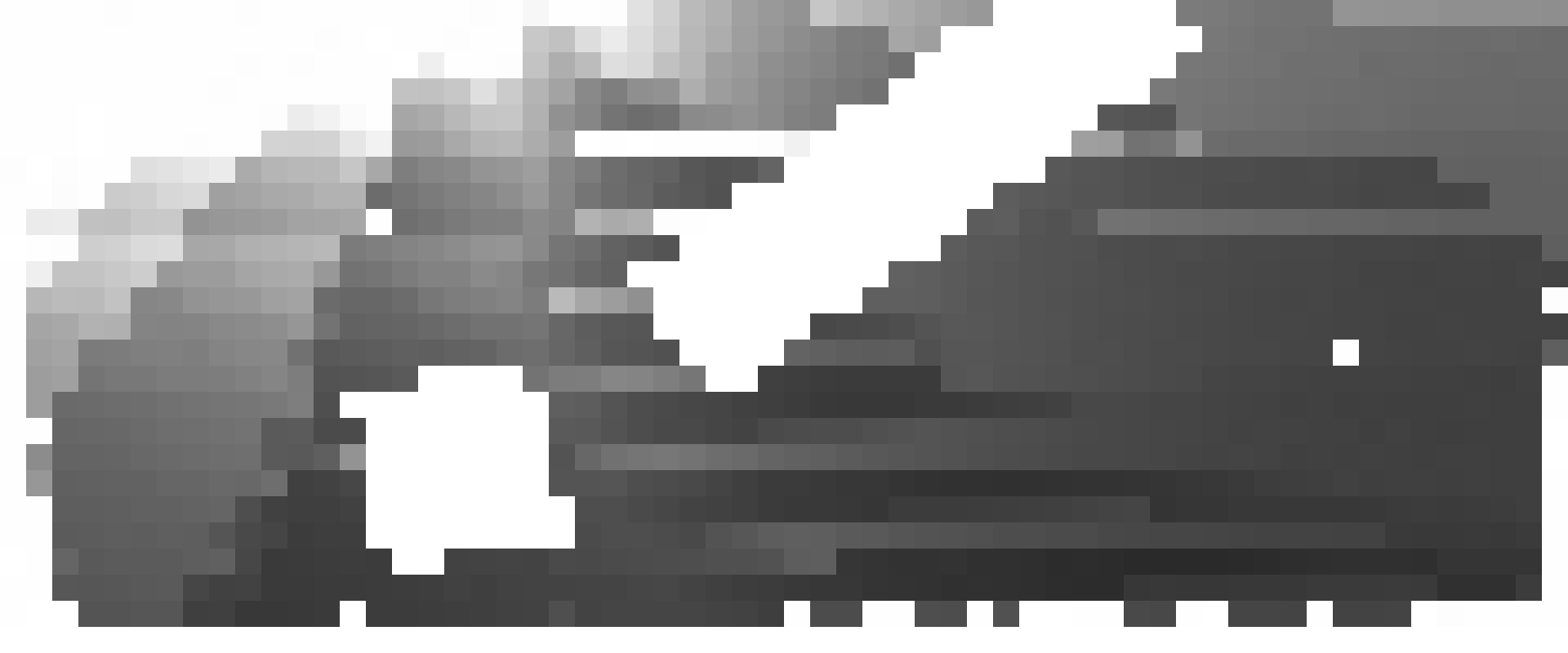}}%
            \hspace{1pt}%
            \fbox{\includegraphics[
                width=\dimexpr0.235\columnwidth-2\fboxrule\relax]
                {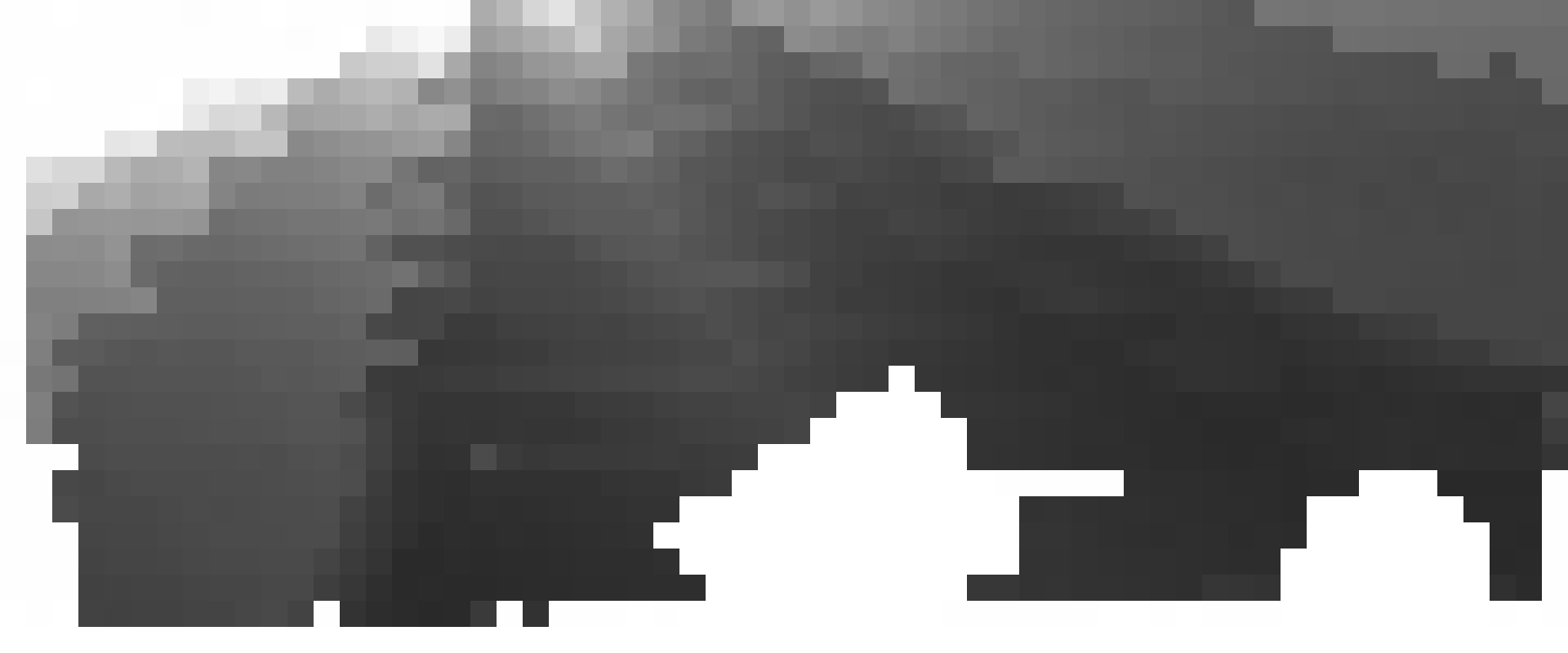}}%
        }
    \end{tabular}
    \caption{Random occlusion masks applied to LiDAR-derived depth images during training at time steps $t$ and $t+1$.}
    \label{fig:depth_corruption}
    \vspace{-2mm}
    \end{figure}
                
    \section{Joint-Embedding Predictive Learning for Perceptive Legged Locomotion}\label{sec:method}

    \subsection{Proprio-Exteroceptive JEPA (PE-JEPA) World Modeling} \label{sec:method_jepa}
    As shown in Fig.~\ref{fig:pipeline}, we introduce modality-specific encoders, GRU memory, and two prediction heads within the JEPA architecture for legged perception, enabling joint prediction of future proprioceptive and LiDAR-derived depth embeddings. We collect proprioceptive observations $\bm{o}_t$ and depth observations $\bm{d}_t$ over windows of length $H_\text{o}$ and $H_\text{d}$, respectively, forming the histories $\bm{O}_t\coloneqq\bm{o}_{t-H_\text{o}+1:t}$ and $\bm{D}_t\coloneqq\bm{d}_{t-H_\text{d}+1:t}$~\cite{luo2024pie}. We similarly define the action history as $\bm{A}_t\coloneqq\bm{a}_{t-H_\text{a}+1:t}$. Given $\bm{O}_t$, $\bm{D}_t$, and $\bm{A}_t$, we have
    \begin{equation*}
        \begin{aligned}
            \text{Proprioception Encoder:}& \quad \bm{z}_t^\text{o} = \operatorname{MLP} \big( \bm{O}_t \big)\,, \\
            \text{Depth Encoder:}& \quad \bm{z}_t^\text{d} = \operatorname{ViT} \big( \bm{D}_t \big)\,, \\
            \text{Memory:}& \quad \bm{h}_t = \operatorname{GRU} \big( \bm{z}_t^\text{o}, \bm{z}_t^\text{d}, \bm{h}_{t-1} \big)\,, \\
            \text{Proprioception Pred. Head:}& \quad \hat{\bm{z}}_{t+1}^\text{o} = \operatorname{MLP} \big( \bm{h}_t, \bm{A}_t \big)\,, \\
            \text{Depth Pred. Head:}& \quad \hat{\bm{z}}_{t+1}^\text{d} = \operatorname{MLP} \big( \bm{h}_t, \bm{A}_t \big)\,.
    	\end{aligned}
    \end{equation*}
    Here, the proprioceptive, exteroceptive, and action history lengths are $H_\text{o}=10$, $H_\text{d}=2$, and $H_\text{a}=5$, respectively.
    
    \subsubsection*{Prediction loss} We train the proposed PE-JEPA world model using a single-step latent prediction loss defined as
    \begin{equation}\label{eq:pred}
        \begin{aligned}
            \mathcal{L}_\text{pred}=&\frac{1}{(T_\text{JEPA}-1) N_\text{env}}
            \lsft{-12mu}\sum_{t=1}^{T_\text{JEPA}-1}\sum_{n=1}^{N_\text{env}}
            \big(\Vert\hat{\bm{z}}_{t+1,n}^\text{o}-\bm{z}_{t+1,n}^\text{o}\Vert^2\\
            &+\Vert \hat{\bm{z}}_{t+1,n}^\text{d} - \bm{z}_{t+1,n}^\text{d} \Vert^2 \big)\,.
        \end{aligned}
    \end{equation}
    Here, $\Vert\cdot\Vert$ denotes the Euclidean norm, $T_\text{JEPA}$ is the JEPA rollout length, and $N_\text{env}=4096$ is the number of parallel environments indexed by $n$.

    \subsubsection*{Sketched Isotropic Gaussian Regularization (SIGReg) loss} The prediction loss alone allows trivially predictable latent embeddings (e.g., constant values). To prevent representation collapse, we also apply the SIGReg loss to the proprioception and depth embeddings~\cite{balestriero2025lejepa,maes2026leworldmodel}. For the proprioceptive and exteroceptive latents $\bm{z}_t^\text{o}$ and $\bm{z}_t^\text{d}\in\R^l$, with $l\in\{32,64\}$, we construct a rollout tensor for either embedding as $\bm{Z}\in\mathbb{R}^{T_\text{JEPA}\times N_\text{env}\times l}$. Its slice at time step $t$ is denoted by $\bm{Z}_t\in\mathbb{R}^{N_\text{env}\times l}$. Further, we randomly sample $M=512$ projection directions $\bm{u}^{(1)},\dots,\bm{u}^{(M)}\in\mathbb{S}^{l-1}$ on the unit hypersphere and project the embedding slice onto each direction, yielding
    \begin{equation*}
        \bm{q}_t^{(m)} = \bm{Z}_t \bm{u}^{(m)} \in \mathbb{R}^{N_\text{env}}\,, \eqwith m=1,\dots,M\,.
    \end{equation*}
    This generates 1-D projections of the original embeddings along $\bm{u}^{(m)}$. We evaluate the Epps-Pulley test statistic $\mathcal{T}(\cdot)$ on the projected samples to measure their deviation from the standard Gaussian distribution $\mathcal{N}(0,1)$~\cite{epps1983test}. Minimizing this statistic across the sampled directions encourages the joint embedding distribution to approach an isotropic Gaussian~\cite{cramer1936some}. Therefore, we optimize the sum of Epps-Pulley statistics over all $M$ directions. Treating the time dimension $T_\text{JEPA}$ as the batch dimension, we define our SIGReg loss as
    \begin{equation*}
        \text{SIGReg} \big(\bm{Z}\big)=\frac{1}{T_\text{JEPA}\cdot M} \sum_{t=1}^{T_\text{JEPA}} \sum_{m=1}^M \mathcal{T} \big( \bm{q}_t^{(m)} \big)\,.
    \end{equation*}

    \subsubsection*{Final loss} We apply SIGReg to both the proprioceptive embeddings $\bm{Z}^\text{o}$ and LiDAR-derived exteroceptive embeddings $\bm{Z}^\text{d}$ collected during rollout. Combining the prediction and SIGReg losses, the final JEPA loss is
    \begin{equation}\label{eq:jepa}
        \mathcal{L}_\text{JEPA} = \mathcal{L}_\text{pred} + \lambda \big( \text{SIGReg} (\bm{Z}^\text{o}) + \text{SIGReg} (\bm{Z}^\text{d}) \big)\,.
    \end{equation}
    The SIGReg weight $\lambda=0.1$. By jointly learning the encoders and predictor, the model can learn representations that are less sensitive to noise, occlusion, and distortion in depth images derived from irregular LiDAR scans.
    \begin{figure}[t!]
        \vspace{2mm}
        \centering
        \setlength{\fboxsep}{0pt}
        \setlength{\fboxrule}{0.3pt}
        \setlength{\tabcolsep}{1.5pt}

        \begin{tabular}{@{}cccc@{}}
        \fbox{\includegraphics[width=\dimexpr0.235\columnwidth-2\fboxrule\relax]{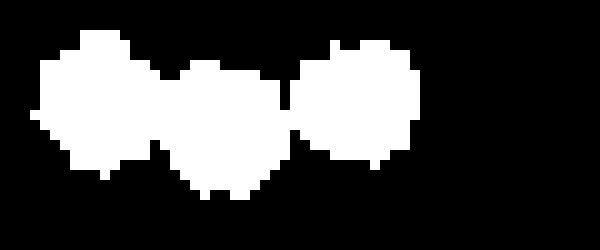}}&
        \fbox{\includegraphics[width=\dimexpr0.235\columnwidth-2\fboxrule\relax]{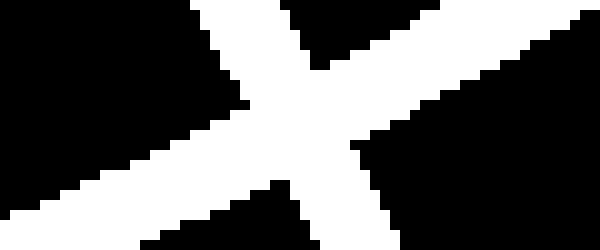}}&
        \fbox{\includegraphics[width=\dimexpr0.235\columnwidth-2\fboxrule\relax]{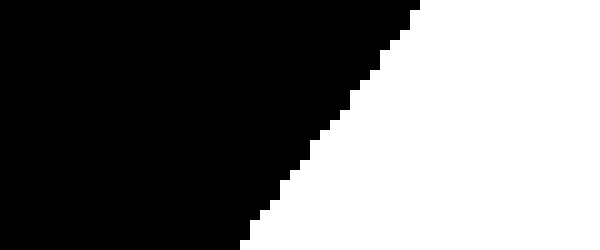}}&
        \fbox{\includegraphics[width=\dimexpr0.235\columnwidth-2\fboxrule\relax]{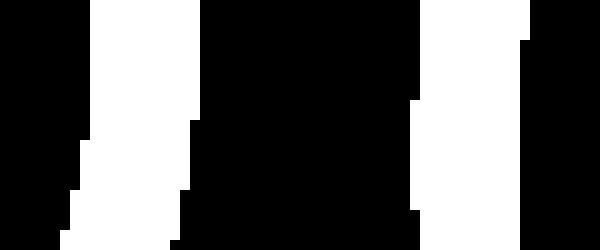}}\\
        (a) \ttt{Blob} & (b) \ttt{Cross} & (c) \ttt{Cut} & (d) \ttt{Stripe}\\[2pt]
        \fbox{\includegraphics[width=\dimexpr0.235\columnwidth-2\fboxrule\relax]{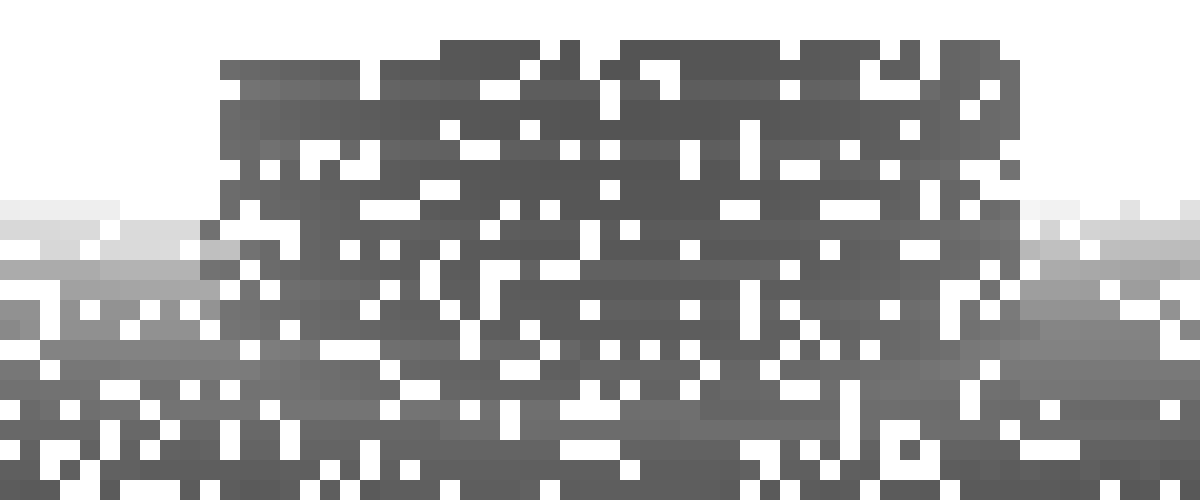}}&
        \fbox{\includegraphics[width=\dimexpr0.235\columnwidth-2\fboxrule\relax]{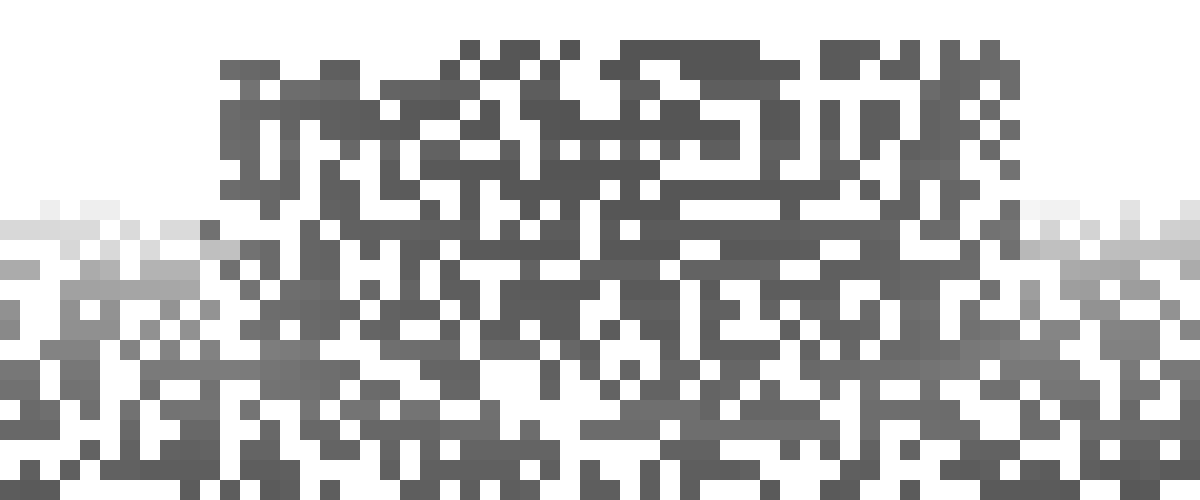}}&
        \fbox{\includegraphics[width=\dimexpr0.235\columnwidth-2\fboxrule\relax]{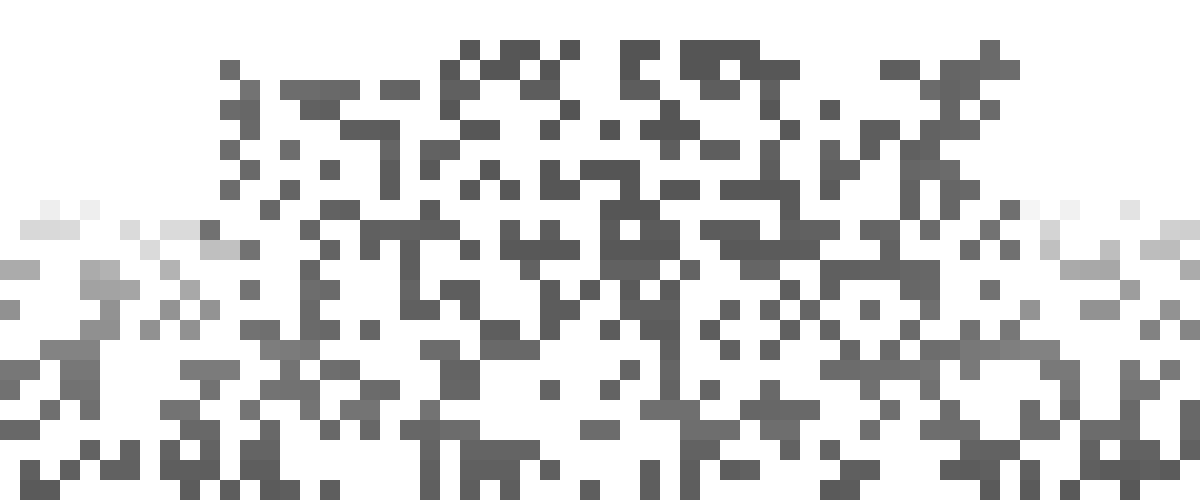}}&
        \fbox{\includegraphics[width=\dimexpr0.235\columnwidth-2\fboxrule\relax]{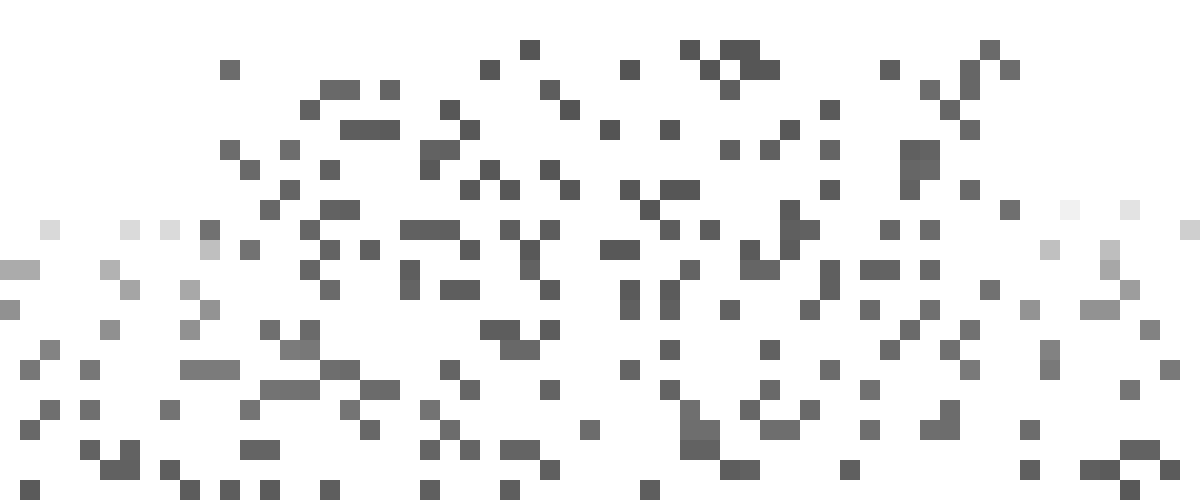}}\\
        \multicolumn{4}{c}{(e) Masks of ratio $20\%$, $40\%$, $60\%$, and $80\%$.}
        \end{tabular}
        \caption{LiDAR occlusion and mask patterns in evaluation.}
        \label{fig:ablation_masks_dropout}
        % \vspace{-2mm}
    \end{figure}

    \begin{figure}[t!]
        % \vspace{2mm}
        \centering
        \includegraphics[width=0.95\linewidth]{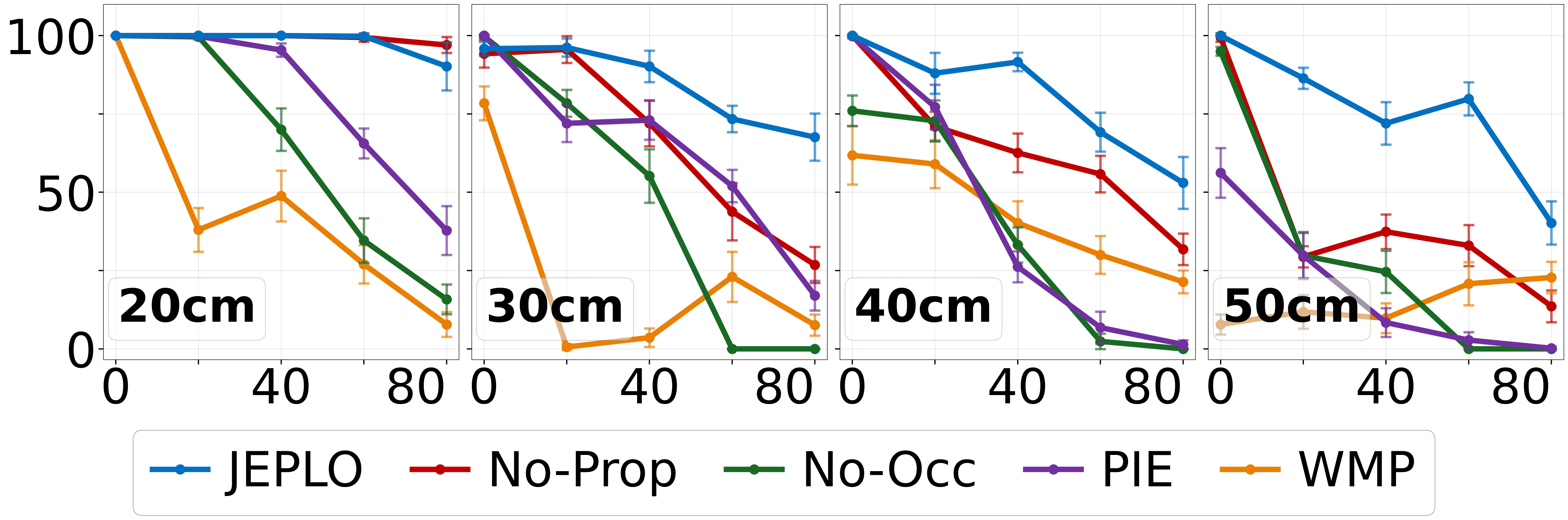}
        \caption{Success rates across mask ratios for box heights.}     
        \label{fig:box_height_masking}
        \vspace{-2mm}
    \end{figure}
    
    \subsection{Concurrent JEPA-Teacher-Student (CJTS) Training}
    Inspired by~\cite{wang2024cts} in blind locomotion, we propose a concurrent pipeline to jointly train PE-JEPA, a privileged teacher encoder, and a deployable student encoder, with both encoders providing inputs to a shared actor $\pi$. This follows
    \begin{align*}
        \text{Teacher Encoder:}& \quad \bm{z}_t^\text{e} = \operatorname{MLP} \big( \bm{O}_t, \bm{m}_t, \bm{s}_t \big)\,, \\
        \text{Student Encoder:}& \quad \hat{\bm{z}}_t^\text{e} = \operatorname{MLP} \big( \bm{O}_t, \operatorname{sg} ( \bm{h}_t) \big)\,, \\
        \text{Actor:}& \quad \bm{a}_t =
        \begin{cases}
            \pi \big( \bm{o}_t, \bm{z}_t^\text{e} \big)\,, & \text{teacher env.} \\
            \pi \big( \bm{o}_t, \operatorname{sg} (\hat{\bm{z}}_t^\text{e})\big)\,, & \text{student env.}
        \end{cases} \\
        \text{Critic:}& \quad V \big( \bm{O}_t, \bm{m}_t, \bm{s}_t \big)\,.
    \end{align*}
    Here, $\operatorname{sg}$ denotes the stop-gradient operator. The teacher embedding $\bm{z}_t^\text{e}$ is computed from the proprioceptive history $\bm{O}_t$, local height scan $\bm{m}_t$, and privileged state $\bm{s}_t$. The student embedding $\hat{\bm{z}}_t^\text{e}$ is computed from inputs available at deployment, namely the JEPA hidden state $\bm{h}_t$ and the same proprioceptive history, and is supervised by the teacher embedding. We train the actor, critic, and teacher encoder using standard PPO~\cite{schulman2017proximal}. In student environments, the student encoder learns to imitate the teacher embedding by minimizing the MSE loss given by
    \begin{equation*}
        \mathcal{L}_\text{s} = \frac{1}{T\cdot N_\text{env}^\text{s}}
        \sum_{t=1}^{T} \sum_{n=1}^{N_\text{env}^\text{s}}
        \big\Vert \hat{\bm{z}}_{t,n}^\text{e} - \operatorname{sg} ( \bm{z}_{t,n}^\text{e}) \big\Vert^2\,.
    \end{equation*}
    Here, $T$ denotes the policy rollout length. The student embedding $\hat{\bm{z}}_t^\text{e}$ is detached at the actor input to prevent PPO gradients from reaching the student encoder. During CTS training, JEPA is optimized solely through $\mathcal{L}_\text{JEPA}$ in \eqref{eq:jepa}, and its hidden state is detached at the student encoder input, preventing task gradients from affecting JEPA. At deployment, we discard the teacher encoder and use only the student encoder, with exteroceptive terrain information provided entirely through the JEPA hidden state $\bm{h}_t$.
        
    \subsubsection*{Rewards} Within the JEPA-informed teacher-student pipeline, we use a simple reward formulation, listed in Tab.~\ref{tab:rewards}, similar to that of prior work for legged blind locomotion~\cite{wu2026toward}. It comprises velocity tracking rewards, regularization, and other common terms, including a foot regulation reward that encourages foot lifting. Compared with previous methods~\cite{cheng2024extreme,lai2025wmp}, our formulation includes neither terrain-specific foot placement terms nor motion references.
            
    \begin{table}[htbp]
        \vspace{2mm}
        \centering
		\caption{Reward functions and weights.}
    	\setlength{\tabcolsep}{1pt}
		\renewcommand{\arraystretch}{1.20}%
		\begin{tabular}{@{}l@{}}
    		\toprule
			\hbox to \columnwidth{\textbf{Reward}\hfil\textbf{Weight}} \\
			\midrule
			\hbox to \columnwidth{Linear velocity tracking:\enspace $\smash{\exp\{-4\|\bm{v}^{\text{cmd}}_{\text{xy}}-\bm{v}_{\text{xy}}\|^{2}\}}$\hfil $2.0$} \\
			\rowcolor{tablerowgray}[0pt][0pt]
			\hbox to \columnwidth{Angular velocity tracking:\enspace $\smash{\exp\{-4(\omega^{\text{cmd}}_{\text{z}} - \omega_{\text{z}})^{2}\}}$\hfil $1.0$} \\
			\hbox to \columnwidth{Linear velocity ($\text{z}$):\enspace $\smash{v_{\text{z}}^{2}}$\hfil $-2\times(1-\min(k/1500,1))$} \\
			\rowcolor{tablerowgray}[0pt][0pt]
			\hbox to \columnwidth{Angular velocity ($\text{xy}$):\enspace $\smash{\|\bm{\omega}_{\text{xy}}\|^{2}}$\hfil $-0.05$} \\
			\hbox to \columnwidth{Joint acceleration:\enspace $\smash{\|\ddot{\bm{\theta}}\|^{2}}$\hfil $-1\times10^{-7}$} \\
			\rowcolor{tablerowgray}[0pt][0pt]
			\hbox to \columnwidth{Joint power:\enspace $\smash{|\bm{\tau}|^\top |\dot{\bm{\theta}}|}$\hfil $-2\times10^{-5}$} \\
			\hbox to \columnwidth{Joint torques:\enspace $\smash{\|\bm{\tau}\|^{2}}$\hfil $-1\times10^{-4}$} \\
			\rowcolor{tablerowgray}[0pt][0pt]
			\hbox to \columnwidth{Base height:\enspace $\smash{(h_\text{b}-h_\text{b}^{*})^2}$\hfil $-1-9\times\min(k/5000,1)$} \\
			\hbox to \columnwidth{Action rate:\enspace $\smash{\|\bm{a}_t - \bm{a}_{t-1}\|^{2}}$\hfil $-0.01$} \\
			\rowcolor{tablerowgray}[0pt][0pt]
			\hbox to \columnwidth{Action smoothness:\enspace $\smash{\|\bm{a}_t - 2\bm{a}_{t-1} + \bm{a}_{t-2}\|^{2}}$\hfil $-0.01$} \\
			\hbox to \columnwidth{Undesired contacts:\enspace $\smash{\sum_{b\in\mathcal{B}_\text{u}}\ind[\|\bm{F}_b\|>\SI{5}{\newton}]}$\hfil $-1.0$} \\
			\rowcolor{tablerowgray}[0pt][0pt]
			\hbox to \columnwidth{Joint position limits:\enspace $\smash{\sum[(\bm{\theta}_{\text{min}}-\bm{\theta})_{+}+(\bm{\theta}-\bm{\theta}_{\text{max}})_{+}]}$\hfil $-2.0$} \\
			\hbox to \columnwidth{Hip position:\enspace $\smash{\sum |\bm{\theta}_{\text{hip}}-\bm{\theta}_{\text{hip,default}}|}$\hfil $-0.05$} \\
			\rowcolor{tablerowgray}[0pt][0pt]
			\hbox to \columnwidth{Thigh/calf position:\enspace $\smash{\sum |\bm{\theta}_{\text{leg}}-\bm{\theta}_{\text{leg,default}}|}$\hfil $-0.01$} \\ \addlinespace[2pt]
			\hbox to \columnwidth{Feet regulation:\enspace $\smash{\sum_f\|\bm{v}^{f}_{\text{xy}}\|^2\exp\{-h_f/(0.025h_\text{b}^{*})\}}$\hfil $-0.05$} \\
    		\bottomrule
		\end{tabular}
        \parbox{\columnwidth}{%
            \footnotesize
            \vspace{4pt}
            Symbol explanation: $h_\mathrm{b}$ denotes the base height, with target $h_\mathrm{b}^{*}=\SI{0.38}{\meter}$. For foot $f$, $h_f$ and $\bm{v}^{f}_{\mathrm{xy}}$ denote its height and planar velocity, respectively. $\bm{F}_b$ denotes the contact-force vector for body $b$, and $\mathcal{B}_\mathrm{u}$ is the set of bodies for which contact is undesired. $\ind[\cdot]$ denotes the indicator function. Positive-part operator $(\cdot)+$ acts elementwise. $k$ indicates the training iteration index.
        }
		\label{tab:rewards}
    \end{table}

    \begin{figure*}[t!]
        \vspace{2mm}
        \centering
        \begin{tabular}{c}
            \includegraphics[width=0.99\textwidth]{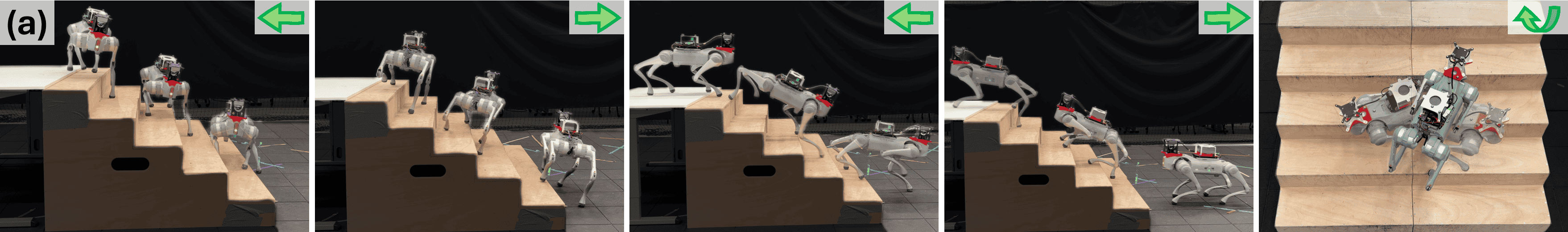}\\
            \includegraphics[width=0.99\textwidth]{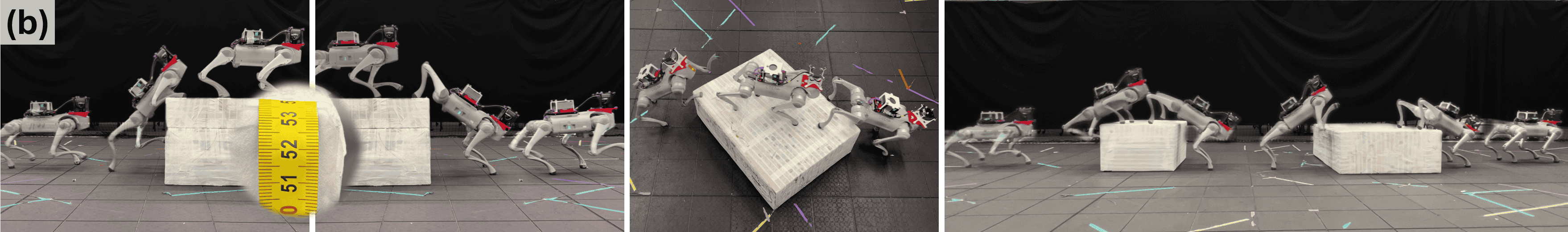}
            \end{tabular}
        \caption{Locomotion across diverse terrains. (a) Omnidirectional stair traversal and in-place rotation on stairs with a \SI{15}{\centi\meter} rise and \SI{25}{\centi\meter} run. (b) Traversal of boxes up to \SI{53}{\centi\meter} high, including angled approaches and consecutive climbs.}
    \label{fig:difficult_tasks}
    \vspace{-2mm}
    \end{figure*}

    \subsection{Training setup} \label{subsec:training}
    We train the policy on the terrains shown in Fig.~\ref{fig:training_terrains}. For \ttt{Rough}, \ttt{Discrete}, \ttt{Pyramid}, and \ttt{InvPyramid} (inverted pyramid), we sample planar and yaw velocity commands independently. For \ttt{Boxes} and \ttt{Steps}, terrain-dependent waypoints determine the body-frame direction of $\bm{v}^{\text{cmd}}_{\text{xy}}$, while the heading error relative to the next waypoint determines $\omega^{\text{cmd}}_{\text{z}}$. During command resampling, we also randomly activate an in-place rotation mode to expose the policy to diverse rotational scenarios. More detailed command-generation parameters are summarized in Tab.~\ref{tab:commands}.
    
    We further apply domain randomization and sensor perturbations using the parameters summarized in Tab.~\ref{tab:domain_randomization}. Additionally, we independently sample 0 to 2 geometric occlusion masks (circles and rectangles) for each LiDAR-derived depth image randomly (Fig.~\ref{fig:depth_corruption}), making these corruptions temporally independent and unpredictable from preceding frames. Under the latent prediction loss $\mathcal{L}_\text{pred}$ in \eqref{eq:pred}, encoding such transient corruptions can increase prediction error, encouraging the encoder to preserve persistent scene geometry while suppressing corruption-specific features.

    As shown in Fig.~\ref{fig:training_terrains}, the teacher's height map $\bm{m}_t\in\mathbb{R}^{300}$ covers a \SI{2.0}{\meter}$\,\times\,$\SI{1.5}{\meter} area centered on the robot at a resolution of \SI{0.1}{\meter}. The privileged state $\bm{s}_t\in\mathbb{R}^{50}$ comprises planar and vertical base velocities $\bm{v}_{\text{xy}}$ and $v_{\text{z}}$, foot heights $h_f$ and contact states, a terrain-type indicator, body-contact force vectors $\bm{F}_b$ indexed by $b$, joint torques $\bm{\tau}$, motor gains, ground friction and restitution, and payload. The policy outputs a 12-D joint-position action $\bm{a}_t$ for the onboard PD controller.

    We allocate $N_\text{env}^\text{s}=1024$ and $N_\text{env}^\text{t}=3072$ environments to the student and teacher, respectively. To stabilize behavior learning together with the proposed PE-JEPA, we gradually introduce student environments through a curriculum. For training iterations $k<5000$, all environments use the teacher branch for rollouts. From $k=5000$ to $k=10000$, we linearly increase $N_\text{env}^\text{s}$ from $0$ to $1024$ and the weight of the student PPO surrogate-loss term from $0$ to $1$. Training is conducted in IsaacLab~\cite{mittal2025isaaclab}, with Livox Mid-360 LiDAR scans simulated following~\cite{yuan2025reasan}. We use the PPO implementation from RSL-RL~\cite{schwarke2025rslrl}, and the complete training process takes about $12$ hours on an NVIDIA RTX 6000 Pro GPU.

    \section{Evaluation in Simulation}\label{sec:eval}
    \subsection{Benchmark setup} 
    To evaluate JEPLO's transfer beyond the training environment, we develop a separate MuJoCo-based environment~\cite{todorov2012mujoco} with simulated LiDAR and perception degradation scenarios. We consider two tasks, climbing a \SI{0.5}{\meter} box and traversing stairs with a rise of \SI{0.2}{\meter} and a run of \SI{0.3}{\meter}, corresponding to the highest terrain difficulty during training. For stair traversal, we challenge the robots with randomly sampled robot headings that require ascent in arbitrary directions. We evaluate four randomized occlusion patterns, shown in Fig.~\ref{fig:ablation_masks_dropout}-(a-d). \ttt{Stripe} resembles occlusion caused by the sensor-cage pillars on the real robot (Fig.~\ref{fig:front}).
    
    We compare JEPLO with two ablations. \textit{No-Prop} removes proprioception from the world model input, while \textit{No-Occ} removes occlusion augmentation during training (Fig.~\ref{fig:depth_corruption}). We also implement two baselines based on state-of-the-art perceptive locomotion approaches from prior work. (1) PIE: we replace PE-JEPA with an estimator from~\cite{luo2024pie} that learns implicit-explicit features, reconstructs terrain height maps, and predicts proprioception. (2) WMP: we replace PE-JEPA with DreamerV3~\cite{hafner2025dreamer} following~\cite{lai2025wmp}, which reconstructs depth and proprioceptive observations and predicts future latent states. Both baselines follow the training setup in Sec.~\ref{subsec:training}. We retain CTS training across all methods to isolate the effect of world modeling and train each for $15000$ iterations.

    \subsection{Results and discussion}
    
    Tab.~\ref{tab:occlusion_success_rates} reports the mean and standard deviation of success rates over $500$ trials per method and occlusion condition for each terrain, \ttt{Stairs} and \ttt{Box}. On stairs, JEPLO achieves the highest success rate under every condition, remaining close to its unoccluded (\ttt{Standard}) performance except under \ttt{Cut}. \textit{No-Prop} is relatively insensitive to occlusion but performs worse overall, suggesting that proprioception contributes to task performance. In contrast, \textit{No-Occ} deteriorates sharply under all occlusion patterns, highlighting the contribution of occlusion augmentation to robustness. PIE and WMP also underperform JEPLO across all conditions, with their largest declines occurring under \ttt{Cut}. For boxes, JEPLO again achieves the highest success rate overall, reaching $100\%$ without occlusion. \textit{No-Prop}, \textit{No-Occ}, and PIE deteriorate sharply under every tested occlusion pattern. Notably, WMP performs worse without occlusion than under any tested occlusion pattern. One possible explanation is that reconstructing random occlusion patterns during training makes it sensitive to the mismatch between clean evaluation inputs and the occlusion-augmented training distribution.
    
    \begin{table}[htbp]
        \centering
        \caption{Benchmark success rates. Bold and underlined values indicate the best and second-best results, respectively.}
        \label{tab:occlusion_success_rates}
        \setlength{\tabcolsep}{1pt}
        \renewcommand{\arraystretch}{1.0}
        \begin{tabularx}{\columnwidth}{l*{5}{>{\centering\arraybackslash}X}}
            \toprule
            \textbf{Occlusion} & JEPLO & \textit{No-Prop} & \textit{No-Occ} & PIE~\cite{luo2024pie} & WMP~\cite{lai2025wmp} \\
            \midrule
            \multicolumn{6}{c}{\textbf{\ttt{Stairs} (rise: \SI[mode=text,reset-text-series=false]{0.2}{\meter}, run: \SI[mode=text,reset-text-series=false]{0.3}{\meter})}} \\
            \midrule
            Standard & $\bestresult{59.2\pm8.3}$ & $34.2\pm6.1$ & $34.2\pm6.5$ & $54.8\pm6.4$ & $\secondresult{57.8\pm6.6}$ \\
            \rowcolor{tablerowgray}
            \ttt{Blob} & $\bestresult{59.8\pm5.6}$ & $36.2\pm7.8$ & $13.0\pm4.1$ & $\secondresult{55.4\pm6.0}$ & $50.2\pm6.2$ \\
            \ttt{Cross} & $\bestresult{60.2\pm5.5}$ & $42.8\pm8.1$ & $14.8\pm4.4$ & $\secondresult{49.6\pm5.4}$ & $41.0\pm7.1$ \\
            \rowcolor{tablerowgray}
            \ttt{Cut} & $\bestresult{39.6\pm7.4}$ & $\secondresult{37.4\pm7.0}$ & $16.4\pm6.1$ & $23.2\pm5.3$ & $31.4\pm6.6$ \\
            \ttt{Stripe} & $\bestresult{62.6\pm8.4}$ & $36.6\pm4.7$ & $11.0\pm3.3$ & $\secondresult{53.6\pm5.9}$ & $52.4\pm6.1$ \\
            \midrule
            \multicolumn{6}{c}{\textbf{\ttt{Box} (height: \SI[mode=text,reset-text-series=false]{0.5}{\meter})}} \\
            \midrule
            Standard & $\bestresult{100\pm0.0}$ & $\secondresult{99.4\pm1.3}$ & $95.0\pm1.4$ & $56.2\pm7.9$ & $7.8\pm3.2$ \\
            \rowcolor{tablerowgray}
            \ttt{Blob} & $\bestresult{57.6\pm9.1}$ & $8.8\pm3.8$ & $12.2\pm5.5$ & $6.0\pm3.5$ & $\secondresult{33.6\pm7.0}$ \\
            \ttt{Cross} & $\bestresult{50.2\pm9.6}$ & $7.4\pm3.4$ & $5.8\pm4.8$ & $4.0\pm1.9$ & $\secondresult{30.8\pm4.7}$ \\
            \rowcolor{tablerowgray}
            \ttt{Cut} & $\bestresult{20.0\pm5.3}$ & $11.4\pm5.6$ & $3.4\pm2.8$ & $1.2\pm1.4$ & $\secondresult{15.0\pm5.1}$ \\
            \ttt{Stripe} & $\bestresult{63.4\pm4.2}$ & $12.4\pm4.8$ & $2.6\pm1.6$ & $4.6\pm3.0$ & $\secondresult{30.8\pm4.8}$ \\
            \bottomrule
        \end{tabularx}
    \end{table}

    \subsubsection*{Discussion} 
    We further evaluate robustness to random scan masking at ratios from $20\%$ to $80\%$ in LiDAR-derived depth images, as depicted in Fig.~\ref{fig:ablation_masks_dropout}-(e), using \ttt{Box} terrains of heights from \SIrange{20}{50}{\cm}. As shown in Fig.~\ref{fig:box_height_masking}, JEPLO exhibits the least performance degradation as the mask ratio increases, with a larger advantage over other methods for higher boxes. This trend is consistent with greater reliance on perception when climbing higher boxes.

    \begin{figure}[t!]
        \vspace{2mm}
        \centering
        \includegraphics[width=0.95\linewidth]{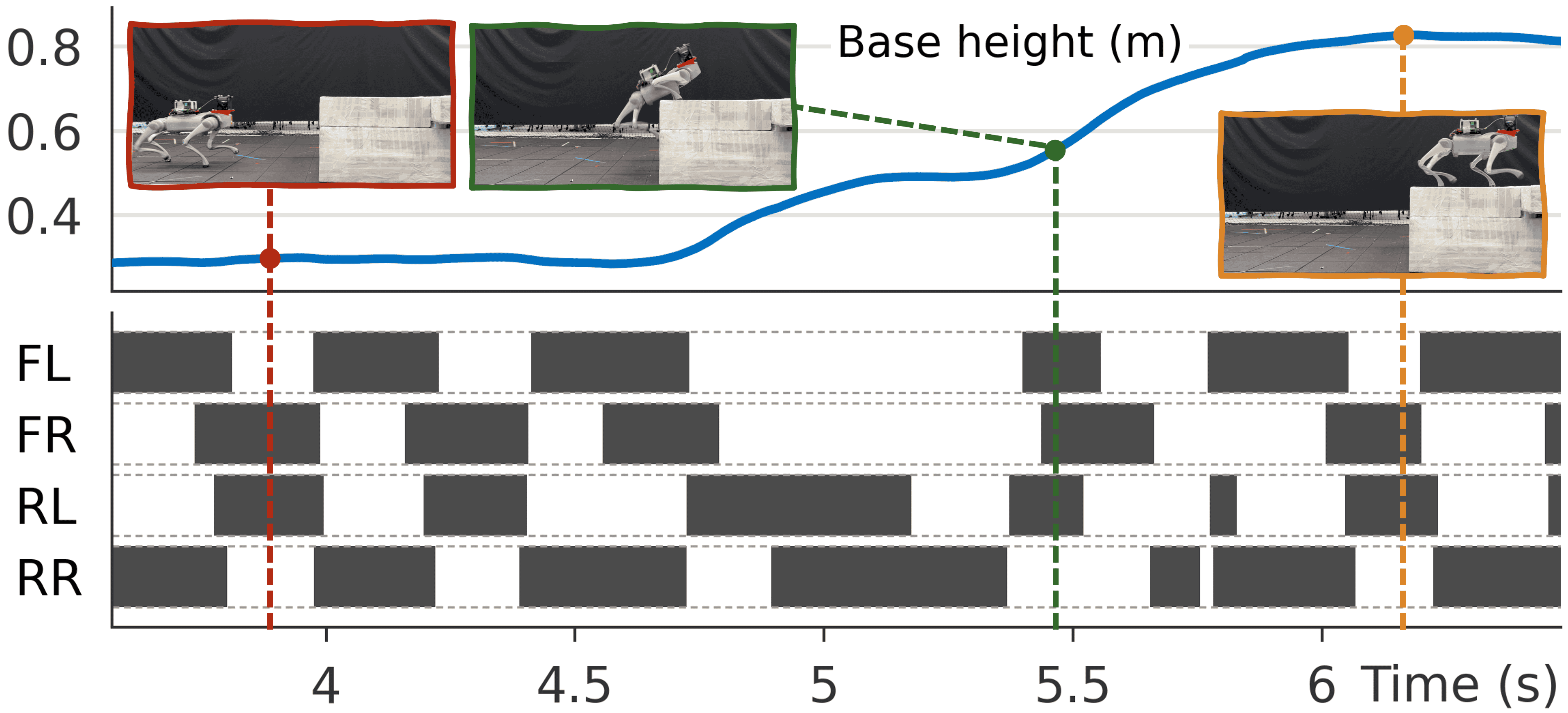}
        \caption{Gait analysis for climbing a \SI{53}{\centi\meter}-high box.}
    \label{fig:gait_analysis}
    \end{figure}
    
    \begin{figure}[t!]
        \centering
        \includegraphics[width=0.95\linewidth]{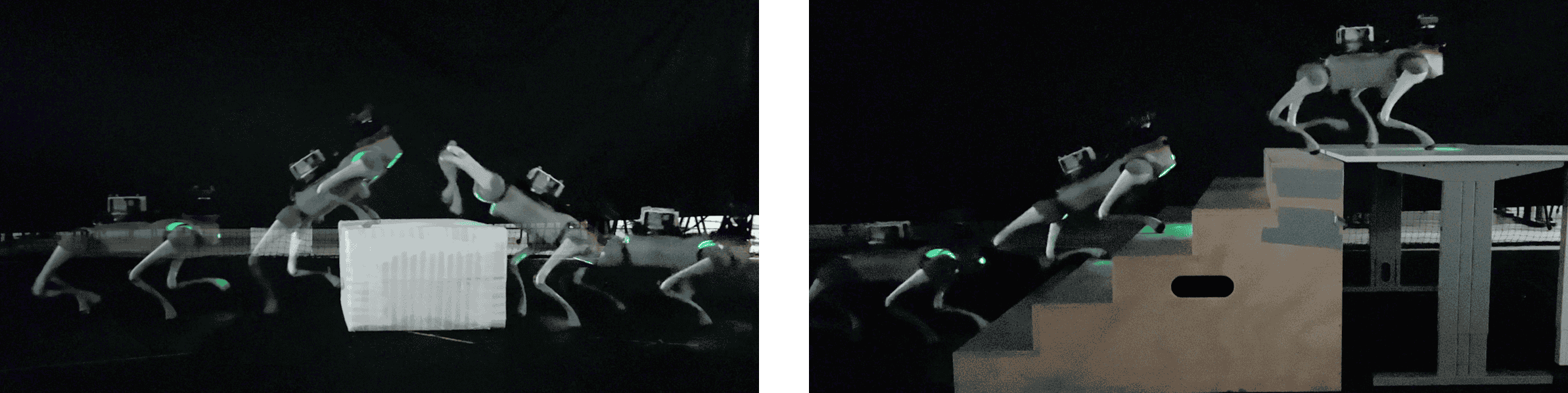}
        \caption{Box and stair climbing in darkness.}
        \label{fig:real_dark}
        \vspace{-2mm}
    \end{figure}

    \section{Real-World Experiments}\label{sec:real_world}
    As shown in Fig.~\ref{fig:front}-(a), we equip a Unitree Go2 quadrupedal robot with an NVIDIA Jetson AGX Orin (\SI{64}{\giga\byte}) and a Livox Mid-360 LiDAR. A perception process receives point clouds from the Livox driver, accumulates typically five successive scans, and projects them into depth images for publication. Our MuJoCo-based evaluation environment uses the same pipeline, enabling seamless sim-to-sim validation and facilitating sim-to-real transfer. A separate control process receives these depth images and controls the robot, running the policy at \SI{50}{\hertz} and the JEPA module at \SI{10}{\hertz}. We use ZeroMQ\footnote{\url{https://zeromq.org}.} for inter-process communication and export the models to ONNX Runtime for TensorRT-accelerated onboard inference. No SLAM or odometry system is required during deployment in velocity tracking. 
        
    As shown in Fig.~\ref{fig:difficult_tasks}-(a), JEPLO enables omnidirectional stair traversal at approximately \SI[per-mode=symbol]{1}{\meter\per\second} and in-place rotation. Fig.~\ref{fig:difficult_tasks}-(b) further demonstrates climbing boxes up to \SI{53}{\centi\meter} high, including consecutive and angled jumps. Fig.~\ref{fig:gait_analysis} visualizes the gait pattern during high-box climbing, showing natural gait coordination and transitions achieved with a simple reward formulation. JEPLO also enables LiDAR-based box and stair climbing in darkness, as shown in Fig.~\ref{fig:real_dark}. In the environment shown in Fig.~\ref{fig:front}-(a), JEPLO completes five consecutive back-and-forth traversals over boxes and up and down stairs without failure. The system also adapts to outdoor environments, including long staircases (over \SI{7.5}{\meter}) and steps of varying geometries. 
        
\begin{figure}[t!]
    \vspace{2mm}
        \centering    
        \begin{tabular}{c}
            \includegraphics[width=0.935\columnwidth]{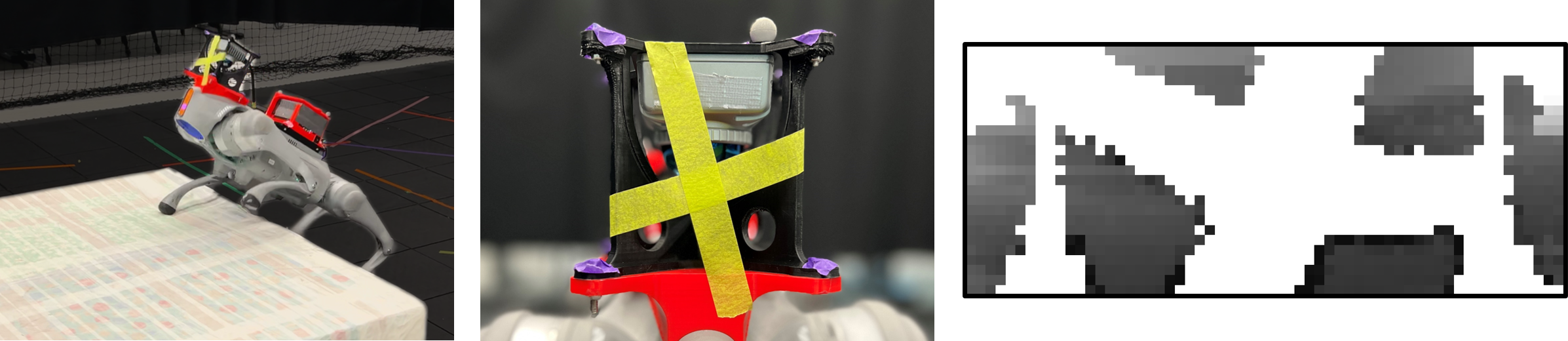}
        \end{tabular}
        \caption{Robust locomotion under LiDAR occlusion.}
        \label{fig:real_world_occlusion}
    \end{figure}

\begin{figure}[t!]
        \centering
        \setlength{\fboxsep}{0pt}
        \setlength{\fboxrule}{0.3pt}
        \setlength{\tabcolsep}{1.5pt}
        \begin{tabular}{@{}ccc@{}}
            \fbox{\adjustbox{trim={0.01\width} {0\height} {0.01\width} {0\height},clip}
            {\includegraphics[width=0.3\columnwidth]{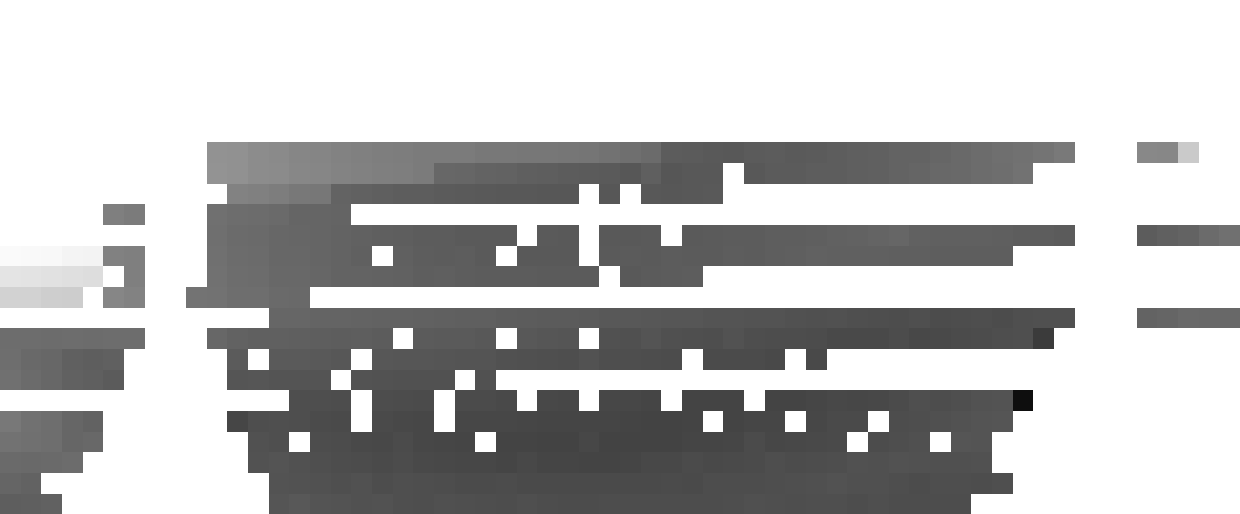}}}&
            \fbox{\adjustbox{trim={0.01\width} {0\height} {0.01\width} {0\height},clip}
            {\includegraphics[width=0.3\columnwidth]{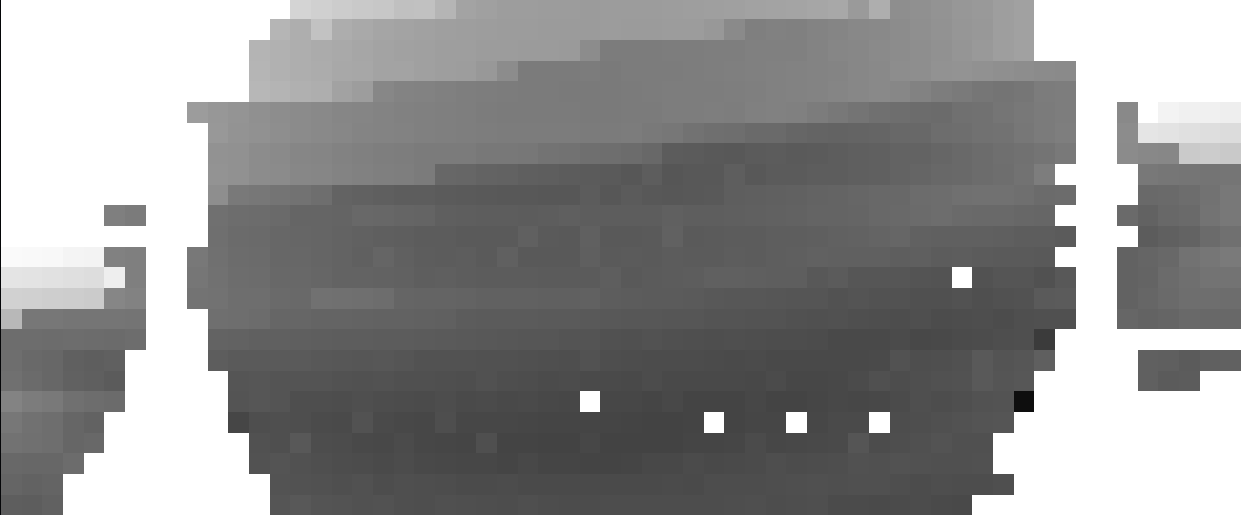}}}&
            \fbox{\adjustbox{trim={0.01\width} {0\height} {0.01\width} {0\height},clip}
            {\includegraphics[width=0.3\columnwidth]{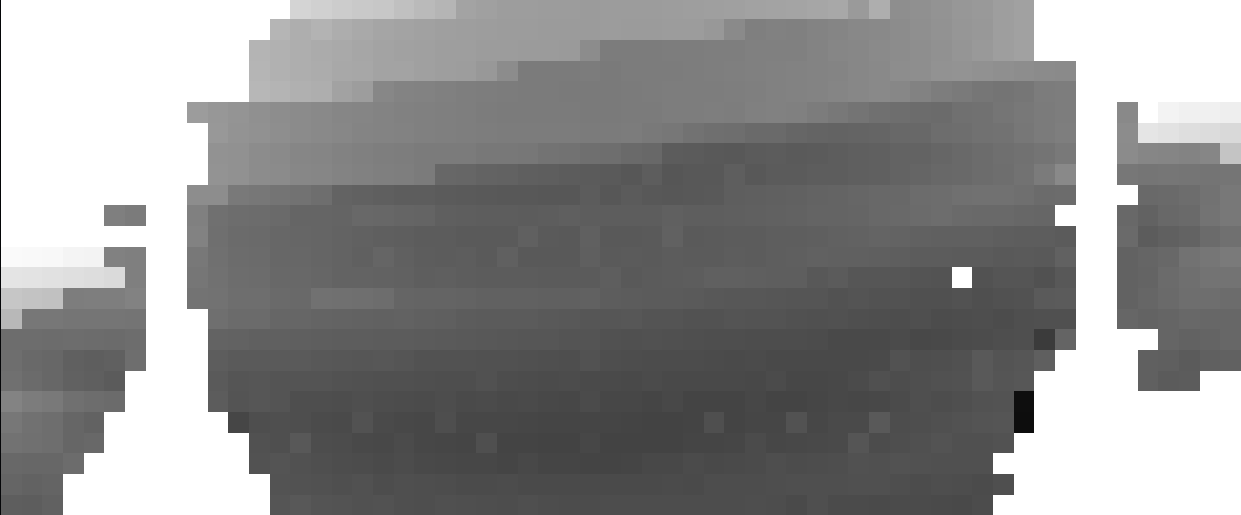}}}\\
            $1$ frame & $3$ frames & $5$ frames
        \end{tabular}
        \caption{Depth images of sparse LiDAR scans.}
        \label{fig:real_sparse}
    \end{figure}

    We further evaluate JEPLO's robustness to real-world sensing degradation on the \SI{34}{\centi\meter}-high box and the stairs shown in Fig.~\ref{fig:difficult_tasks}. For occlusion, we apply cross patterns using tape, as shown in Fig.~\ref{fig:real_world_occlusion}. For sparsity, we reduce the number of accumulated LiDAR frames from the default five to one. Fig.~\ref{fig:real_sparse} compares depth images generated from one, three, and five frames, illustrating how fewer accumulated frames reduce spatial coverage under the non-repetitive scan pattern. The robot successfully traverses both terrain types under these degraded sensing conditions.   
    
    \subsubsection*{Runtime efficiency} 
    We profile JEPLO on the robot's Jetson during a typical perceptive locomotion task, measuring average CPU utilization of $12.4\%$ relative to a single core (12 cores available), memory usage of \SI{1.84}{\giga\byte}, and GPU utilization of $7.35\%$. These results highlight JEPLO's lightweight architecture, enabling onboard perceptive locomotion with low computational and memory demands. 

    \subsubsection*{\ttt{JEPLO} dataset} To support the legged robotics community, we publicly release a dataset recorded during our experiments comprising five indoor sequences and three mixed indoor-outdoor sequences, with complete proprioceptive and exteroceptive recordings. Indoor ground-truth 6-DoF poses are recorded at \SI{100}{\hertz} in a \SI{10}{\meter} $\times$ \SI{4}{\meter} test area using eight Qualisys Miqus M3 motion capture cameras.
    
    \begin{figure*}[t!]
    \vspace{2mm}
        \centering
        \captionsetup[subfigure]{skip=2pt}
        \setlength{\tabcolsep}{2pt}
        \includegraphics[width=0.48\textwidth]{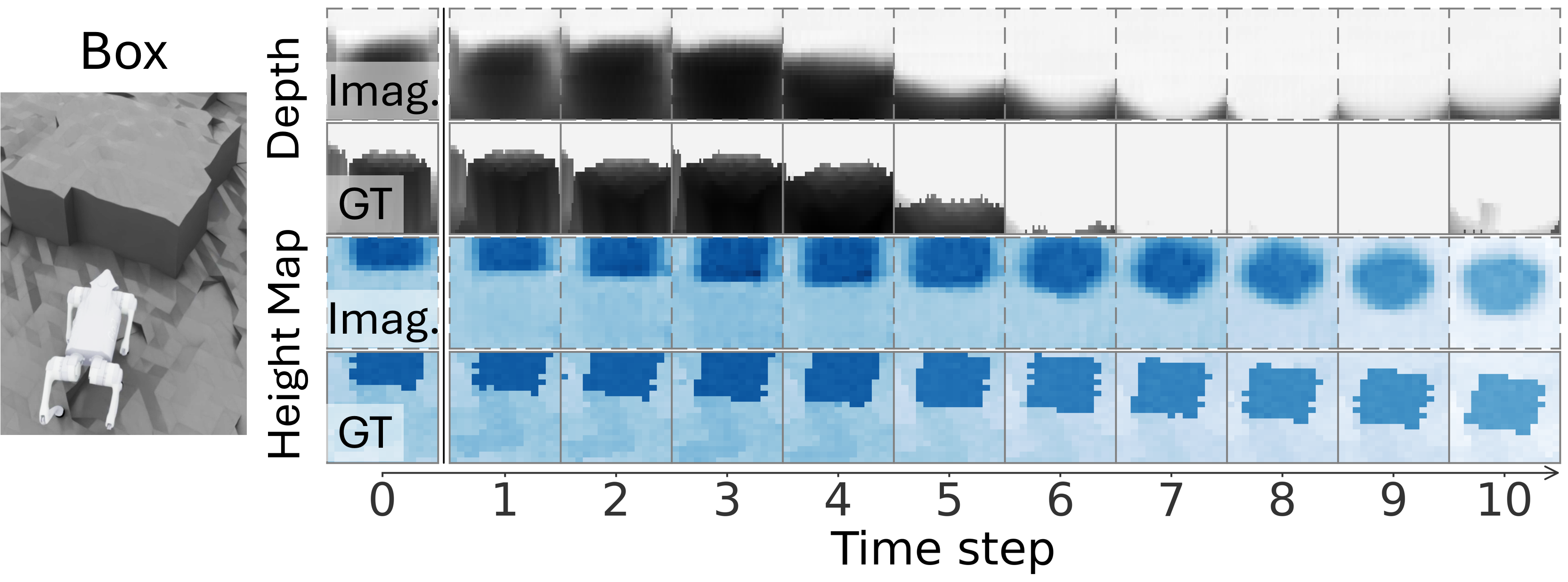}
        \includegraphics[width=0.48\textwidth]{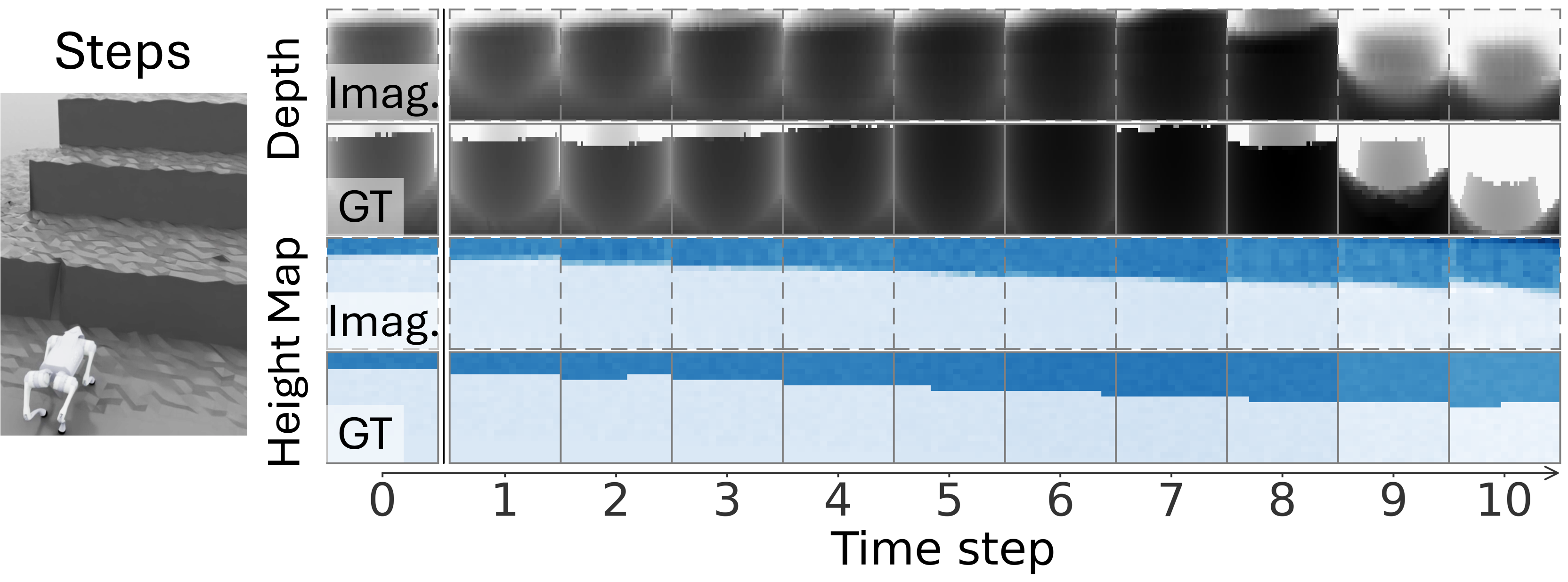}
        \caption{Imagined future depth images and height maps compared with ground truth over \ttt{Box} and \ttt{Steps} terrains.}
        \label{fig:latent_decoding}
        \vspace{-2mm}
    \end{figure*}

    \begin{figure}[t!]
    \centering
    \setlength{\tabcolsep}{0pt}
    \begin{tabular}{cccc}
        \includegraphics[width=0.24\columnwidth]{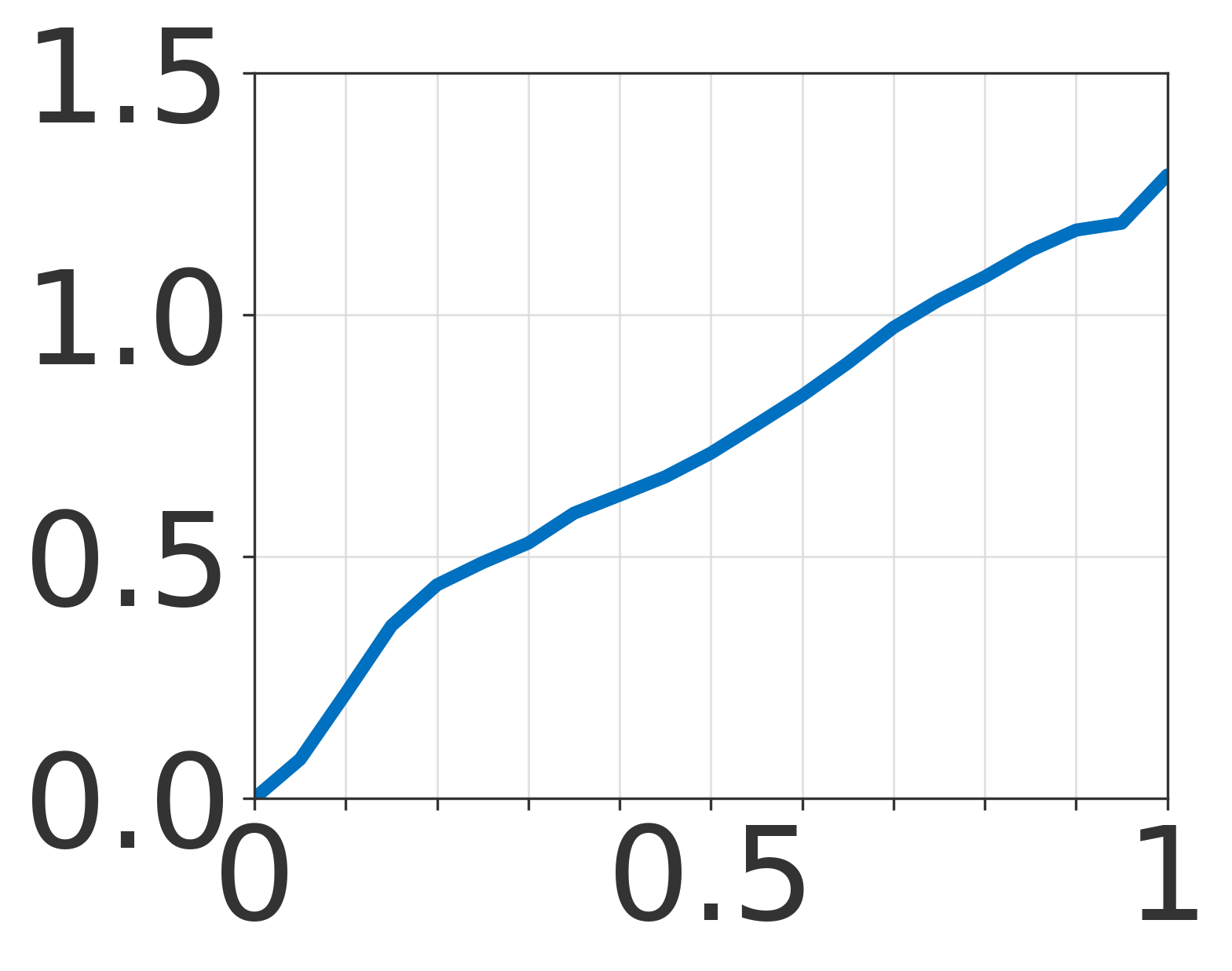} &
        \includegraphics[width=0.24\columnwidth]{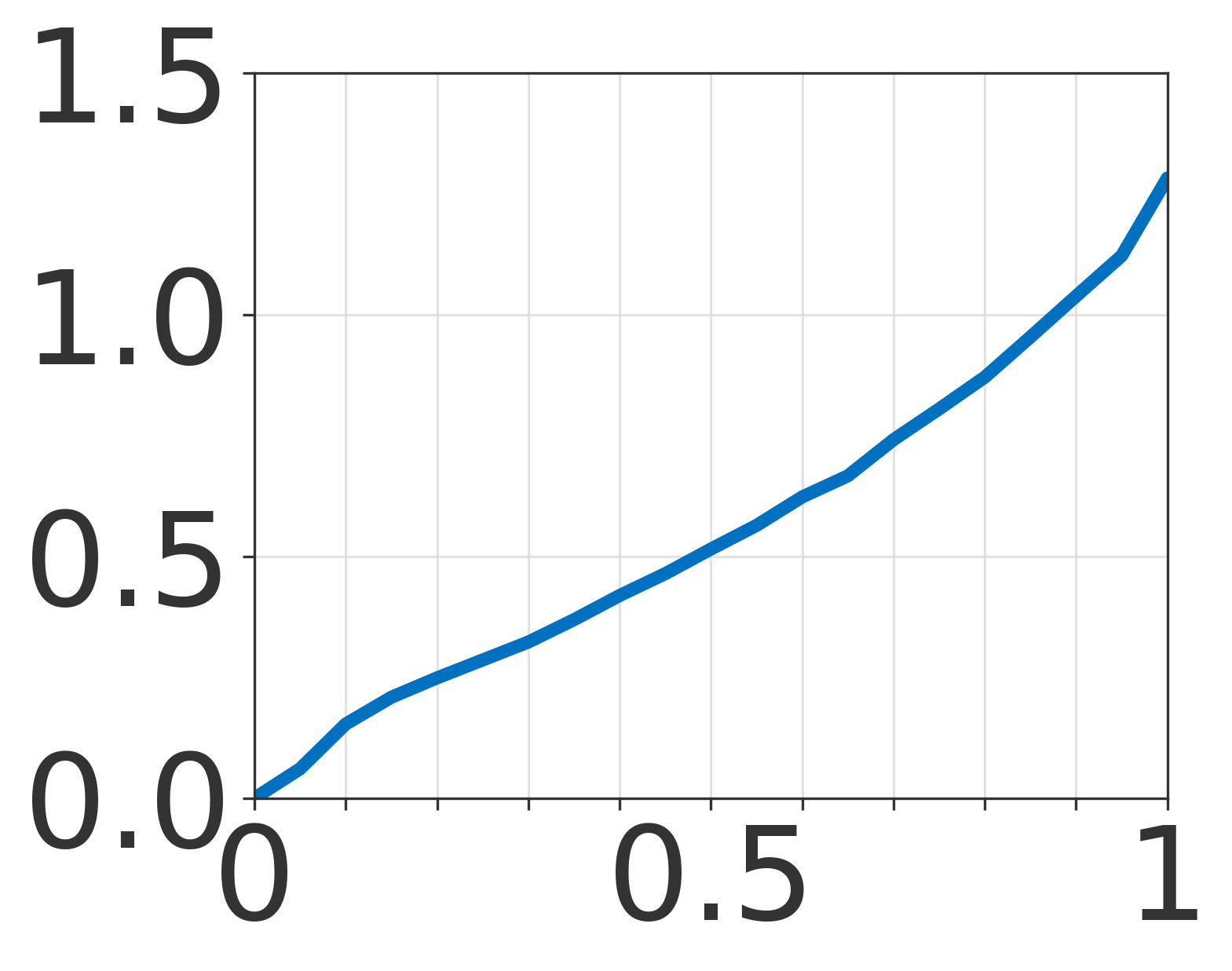} &
        \includegraphics[width=0.24\columnwidth]{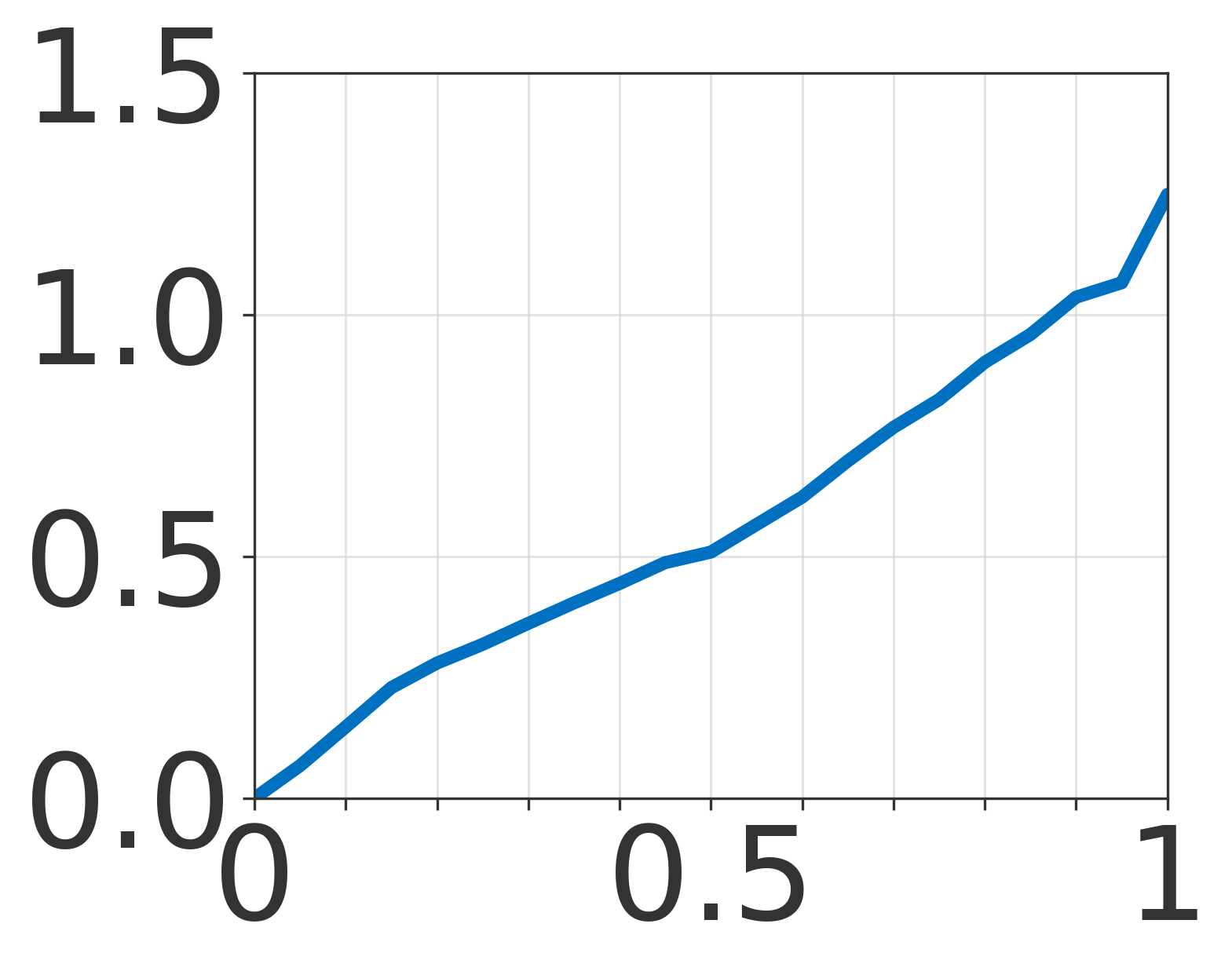} &
        \includegraphics[width=0.24\columnwidth]{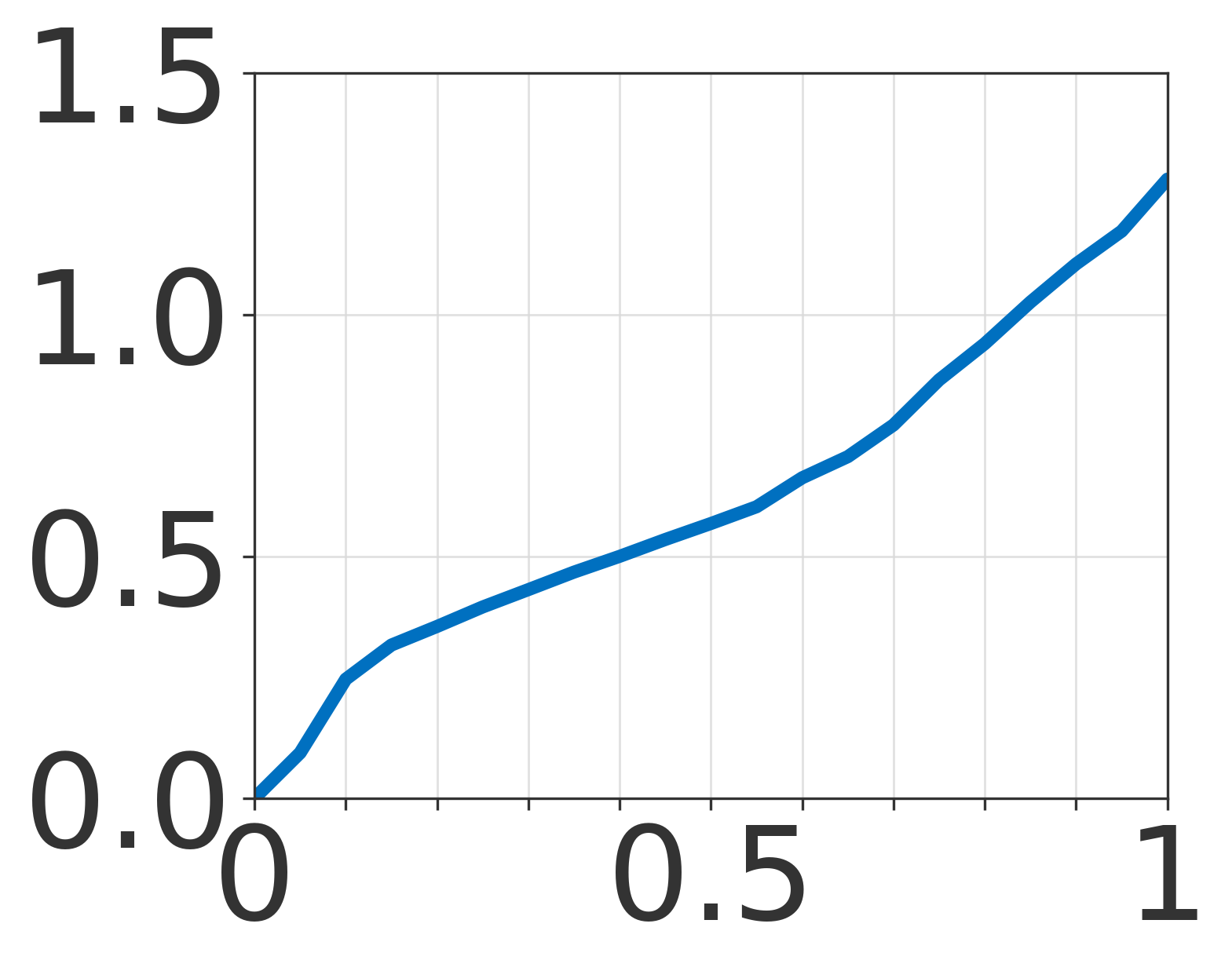}\\
        (a) \ttt{Rough} &(b) \ttt{Steps} &(c) \ttt{Boxes} &(d) \ttt{InvPyr.}    
    \end{tabular}
        \caption{Relative depth latent distances under increasing scan masking over diverse terrains.}
        \label{fig:latent_sweep}
        \vspace{-2mm}
    \end{figure}

    \begin{figure}[t!]
        \centering
        \setlength{\tabcolsep}{0pt}
        \begin{tabular}{cc}
            \includegraphics[width=0.49\columnwidth]{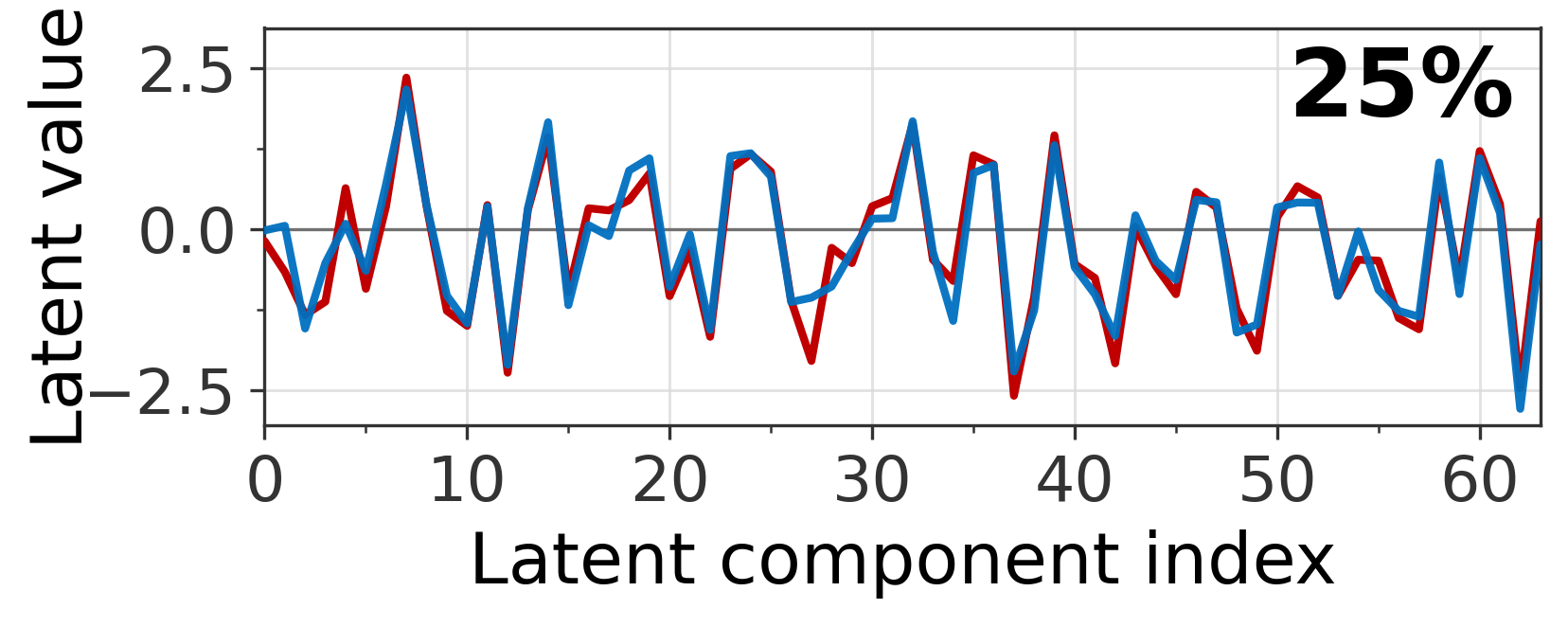} &
            \includegraphics[width=0.49\columnwidth]{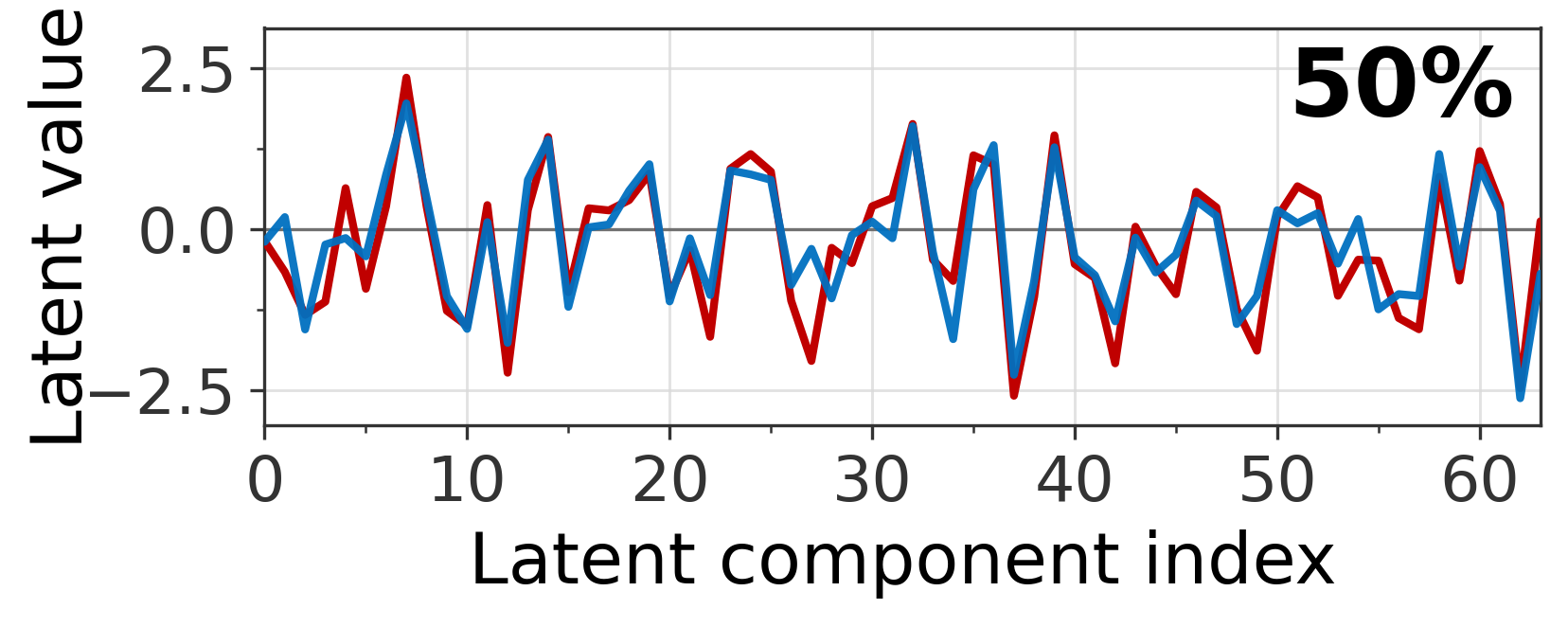} \\
            \includegraphics[width=0.49\columnwidth]{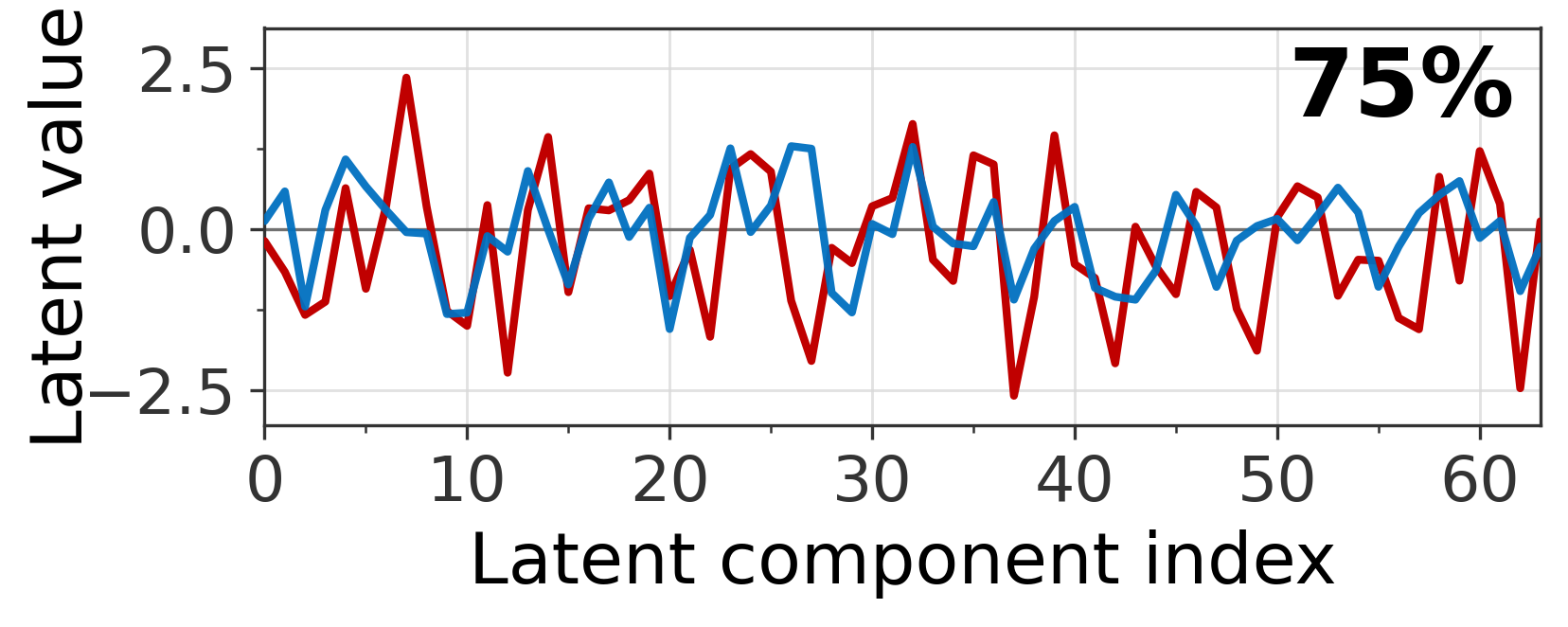} &
            \includegraphics[width=0.49\columnwidth]{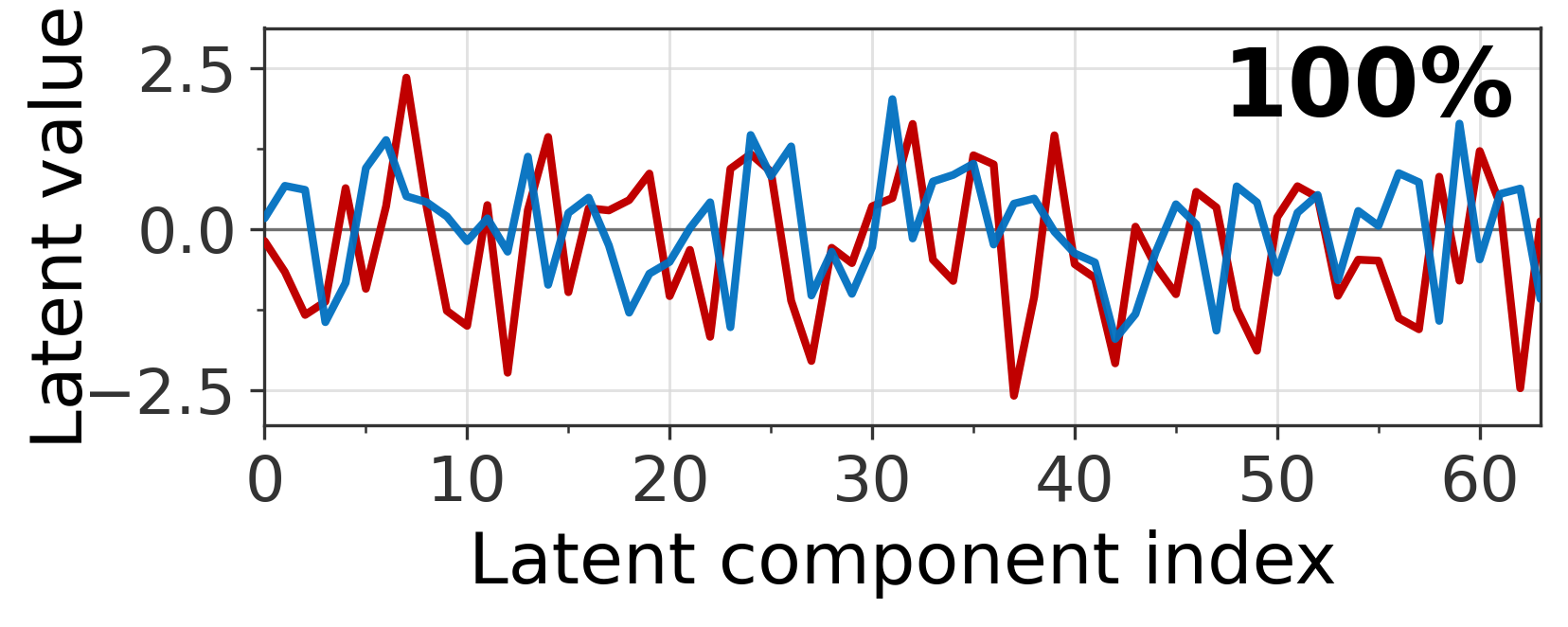} \\
        \end{tabular}
        \caption{Latent representations from unoccluded (red) and $25\%$-$100\%$ masked (blue) scans in \ttt{Steps}.}
        \label{fig:latent_drop_compare}
        \vspace{-3mm}
    \end{figure}

    \section{Latent Analysis}

    To quantify the sensitivity of the learned depth representations to perception degradation, we randomly sample 1000 depth frames from each of \ttt{Rough}, \ttt{Steps}, \ttt{Boxes}, and \ttt{InvPyramid} (Fig.~\ref{fig:training_terrains}). For each frame, we vary the pixel mask ratio $p_{\text{mask}}\in[0,1]$ and compute the relative latent distance between the masked depth latent $\bm{z}^\text{d}(p_{\text{mask}})\in\mathbb{R}^{64}$ and its unoccluded counterpart $\bm{z}^\text{d}(0)$ as $\Vert\bm{z}^\text{d}(p_{\text{mask}})-\bm{z}^\text{d}(0)\Vert/\Vert\bm{z}^\text{d}(0)\Vert$. Fig.~\ref{fig:latent_sweep} shows similar increases in relative latent distance across all four terrains, suggesting that sensitivity to pixel masking is largely consistent across terrain types. For a representative frame, Fig.~\ref{fig:latent_drop_compare} illustrates latent representation variations induced by masking, with pronounced deviations from standard representation at $75\%$.    
    
\begin{figure}[t!]
        \centering
        \setlength{\tabcolsep}{0pt}
        \begin{tabular}{c}
            \includegraphics[width=0.95\columnwidth]{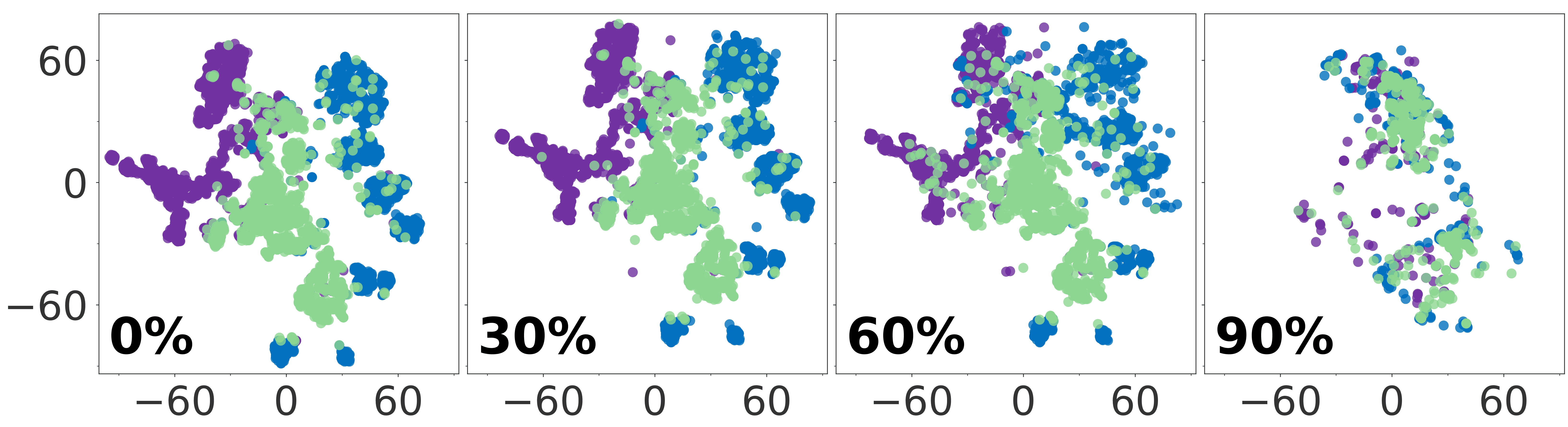} \\
            (a) Exteroceptive latents $\bm{z}_t^\text{d}$ \\
            \includegraphics[width=0.95\columnwidth]{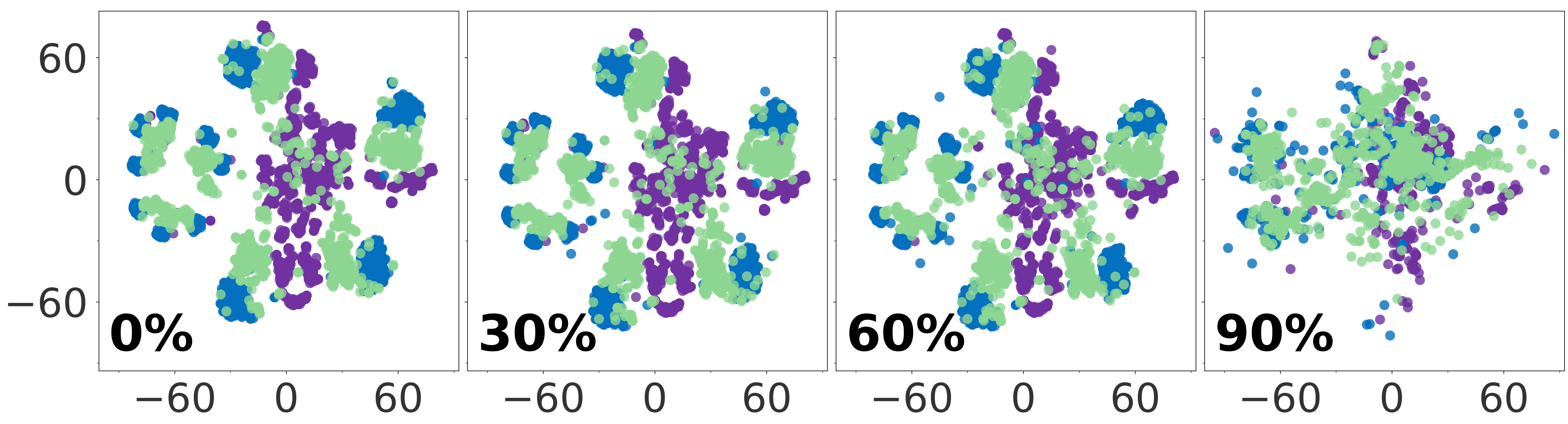} \\
            (b) JEPA hidden states $\bm{h}_t$
        \end{tabular}
        \caption{t-SNE latent visualization under masks for \ttt{Steps} (purple), \ttt{Rough} (blue), and \ttt{Stairs} (green).}
        \label{fig:latent_tsne}
        \vspace{-2mm}
    \end{figure}

    To examine collective latent behavior under perception degradation, we visualize the depth latents $\bm{z}_t^\text{d}$ and JEPA hidden states $\bm{h}_t$ using t-SNE, mapping representations from degraded frames into reference embeddings constructed from clean observations~\cite{opentsne}. At scan masking ratios $p_{\text{mask}}\in\{30\%,60\%,90\%\}$, masked representations are positioned relative to their clean references while the reference coordinates remain fixed. For JEPA hidden states, we apply masking from the start of each episode to account for the GRU's recurrent memory. As shown in Fig.~\ref{fig:latent_tsne}, masked samples largely remain near their clean neighborhoods at $30\%$ and $60\%$ masking, while pronounced deviations emerge at $90\%$. These results provide qualitative evidence of representation robustness under perception degradation.

    \subsubsection*{Latent decoding}
    To examine the task-relevant information retained by the learned representations, we freeze the trained PE-JEPA world model and policy and train two separate Transformer decoders to reconstruct the $20\times15$ local height map from the JEPA hidden state $\bm{h}_t$ and the input depth image from the depth latent $\bm{z}_t^\text{d}$. Each decoder uses a 192-D embedding, three layers, and six attention heads. Starting from the hidden state at time $t$, we roll out the frozen PE-JEPA autoregressively using recorded policy actions and decode the predicted representations over $10$ steps at \SI{0.1}{\second} intervals. As shown by the decoded depth and height maps in Fig.~\ref{fig:latent_decoding}, PE-JEPA accurately predicts climbing timing and terrain interaction without further sensor input, demonstrating its capacity to model task-relevant terrain dynamics.

    \section{Conclusion}\label{sec:conclusion}
    We presented JEPLO, a framework for mapping-free, LiDAR-based perceptive locomotion that combines a proprio-exteroceptive JEPA (PE-JEPA) world model with RL-based policy learning through a single-stage concurrent JEPA-teacher-student (CJTS) pipeline. By predicting in embedding space without explicit observation reconstruction, PE-JEPA learns terrain representations that inform robust locomotion with a simple reward formulation. Simulation and real-world evaluations demonstrated omnidirectional traversal over diverse terrains and resilience to LiDAR occlusion and sparsity. Latent analysis further showed retention of task-relevant terrain information and predictive dynamics under degraded perception. These results support joint-embedding predictive learning as a promising approach to lightweight and resilient legged autonomy.

    \section*{APPENDIX}
    
    \subsection{Training settings and domain randomization}
    We use three command modes. \textit{Direct} mode samples velocity commands for \ttt{Rough}, \ttt{Discrete}, \ttt{Pyramid}, and \ttt{InvPyramid}, together with the in-place \textit{Rotation} mode. \textit{Waypoint} mode guides the robot through successive waypoints on \ttt{Boxes} and \ttt{Steps} to encourage terrain interaction.   Parameters for command generation and domain randomizations are listed in Tab.~\ref{tab:commands} and Tab.~\ref{tab:domain_randomization}, respectively. 
    
    \begin{table}[htbp]
        \centering
        \caption{Parameters for command generation.}
        \setlength{\tabcolsep}{4pt}
        \begin{tabularx}{\columnwidth}{l>{\raggedright\arraybackslash}Xl}
            \toprule
            \textbf{Mode} & \textbf{Parameter} & \textbf{Value} \\
            \midrule
            All & Resampling (\si{\hertz}) & \SI{0.25}{} \\
            \midrule
            \multirow{3}{*}{Direct}
            & $v^{\text{cmd}}_{\text{x}}$ (\si[per-mode=symbol]{\meter\per\second}) & $[-1,-0.5]\cup[0.5,1]$\,, $p_{\text{nz}}=0.9$ \\
            & \cellcolor{tablerowgray}$v^{\text{cmd}}_{\text{y}}$ (\si[per-mode=symbol]{\meter\per\second}) & \cellcolor{tablerowgray}$[-1,-0.5]\cup[0.5,1]$\,, $p_{\text{nz}}=0.75$ \\
            & $\omega^{\text{cmd}}_{\text{z}}$ (\si[per-mode=symbol]{\radian\per\second}) & $[-2,2]$\,, $p_{\text{nz}}=0.5$ \\
            \midrule
            \multirow{3}{*}{Waypoint}
            & Speed (\si[per-mode=symbol]{\meter\per\second}) & $[0.5,1]$\,, $p_{\text{nz}}=0.95$ \\
            & \cellcolor{tablerowgray} Yaw rate clip (\si[per-mode=symbol]{\radian\per\second}) & \cellcolor{tablerowgray}$[-1,1]$ \\
            & Distance threshold (\si{\meter}) & $0.5$ \\
            \midrule
            \multirow{3}{*}{Rotation}
            & Sampling prob. & $0.2$ (iteration $\geq1000$) \\
            & \cellcolor{tablerowgray}\makecell[l]{$\bm{v}^{\text{cmd}}_{\text{xy}}$ (\si[per-mode=symbol]{\meter\per\second})} & \cellcolor{tablerowgray}\mbox{\makecell[l]{$\bm{0}$}} \\
            & \makecell[l]{$\omega^{\text{cmd}}_{\text{z}}$ (\si[per-mode=symbol]{\radian\per\second})} & \mbox{\makecell[l]{$[-2,2]$}} \\
            \bottomrule
        \end{tabularx}%
        \label{tab:commands}
        \parbox{\columnwidth}{%
            \footnotesize
            \vspace{4pt}
            Commands are sampled uniformly over the specified ranges, with $p_{\text{nz}}$ denoting the sampling probability; otherwise, they are set to zero.
        }
        % \vspace{-2mm}
    \end{table}

    \begin{table}[htbp]
        \centering
        \caption{Parameters for domain randomization. }
        \setlength{\tabcolsep}{4pt}
        \begin{tabularx}{\columnwidth}{l>{\raggedright\arraybackslash}Xl}
            \toprule
            \textbf{Domain} & \textbf{Parameter} & \textbf{Range} \\
            \midrule
            \multirow{6}{*}{Dynamics}
            & Mass offset (\si{\kilogram}) & \makecell[l]{Base: $[-2.0,5.0]$} \\
            & \cellcolor{tablerowgray}Base-CoM offset (\si{\meter}) & \cellcolor{tablerowgray}$x,y,z\in[-0.03,0.03]$ \\
            & Friction & $[0,2.0]$ \\
            & \cellcolor{tablerowgray}Restitution & \cellcolor{tablerowgray}$[0,0.5]$ \\
            & Initial joint-pos. scale & $[0.5,1.5]$ \\
            & \cellcolor{tablerowgray}\makecell[l]{Linear \& angular push (\SI{0.25}{\hertz})\\(\si[per-mode=symbol]{\meter\per\second}, \si[per-mode=symbol]{\radian\per\second})} & \cellcolor{tablerowgray}\mbox{\makecell[l]{$\Delta\bm{v}_{\text{xy}}\in[-0.4,0.4]^2$\\$\Delta\bm{\omega}_{\text{xy}}\in[-0.6,0.6]^2$\\$\Delta\omega_{\text{z}}\in[-0.6,0.6]$}} \\
            \midrule
            \multirow{4}{*}{Motor}
            & \makecell[l]{$K_{\text{p}}$ (\si[per-mode=symbol,inter-unit-product=\ensuremath{{}\cdot{}}]{\newton\meter\per\radian})} & $[21,39]$ \\
            & \cellcolor{tablerowgray}\makecell[l]{$K_{\text{d}}$ (\si[per-mode=symbol,inter-unit-product=\ensuremath{{}\cdot{}}]{\newton\meter\second\per\radian})} & \cellcolor{tablerowgray}$[0.7,1.3]$ \\
            & Zero-position offset (\si{\radian}) & $[-0.035,0.035]$ \\
            & \cellcolor{tablerowgray}Action delay (\si{\milli\second}) & \cellcolor{tablerowgray}\mbox{\makecell[l]{$0-20$}} \\
            \midrule
            \multirow{4}{*}{\makecell[l]{Proprio.\\noise}}
            & Angular velocity (\si[per-mode=symbol,inter-unit-product=\ensuremath{{}\cdot{}}]{\radian\per\second}) & $[-0.2,0.2]$ \\
            & \cellcolor{tablerowgray}Projected gravity & \cellcolor{tablerowgray}$[-0.05,0.05]$ \\
            & Joint-position (\si{\radian}) & $[-0.01,0.01]$ \\
            & \cellcolor{tablerowgray}Joint-velocity (\si[per-mode=symbol,inter-unit-product=\ensuremath{{}\cdot{}}]{\radian\per\second}) & \cellcolor{tablerowgray}$[-0.2,0.2]$ \\
            \midrule
            \multirow{3}{*}{LiDAR}
            & Origin drift per axis (\si{\meter}) & $[-0.03,0.03]$ \\
            & \cellcolor{tablerowgray}Range noise (\si{\meter}) & \cellcolor{tablerowgray}$\mathcal{N}(0,0.005^2)$ \\
            & Depth delay (\si{\second}) & $0.1$ \\
            \bottomrule
        \end{tabularx}%
        \label{tab:domain_randomization}
    \end{table}

    \section*{ACKNOWLEDGMENT}
    This paper uses ChatGPT for editing and debugging code.

	\bibliographystyle{IEEEtran.bst}
	\bibliography{bibliography.bib}
	
\end{document}